\let\mypdfximage\pdfximage
\def\pdfximage{\immediate\mypdfximage}

\documentclass[runningheads]{llncs}
\usepackage{graphicx}
\usepackage{amsmath,amssymb} 
\usepackage{color}
\usepackage[width=122mm,left=12mm,paperwidth=146mm,height=193mm,top=12mm,paperheight=217mm]{geometry}
\begin{document}
\pagestyle{headings}
\mainmatter

\newcommand{\repeatthanks}{\textsuperscript{\thefootnote}}

\title{HiDDeN: Hiding Data With Deep Networks} 

\titlerunning{HiDDeN: Hiding Data With Deep Networks}

\authorrunning{Zhu*, Kaplan*, Johnson and Fei-Fei}

\author{Jiren Zhu\thanks{These authors contributed equally.}, Russell Kaplan\repeatthanks, Justin Johnson and Li Fei-Fei}
\newcommand{\etal}{\textit{et al}.}

\institute{Computer Science Department, Stanford University\\
	\email{ \{jirenz,rjkaplan,jcjohns,feifeili\}@cs.stanford.edu}
}

\maketitle

\begin{abstract}
    Recent work has shown that deep neural networks are highly sensitive to tiny perturbations of input images, giving rise to \textit{adversarial examples}. Though this property is usually considered a weakness of learned models, we explore whether it can be beneficial. We find that neural networks can learn to use invisible perturbations to encode a rich amount of useful information. In fact, one can exploit this capability for the task of data hiding. We jointly train encoder and decoder networks, where given an input message and cover image, the encoder produces a visually indistinguishable encoded image, from which the decoder can recover the original message. We show that these encodings are competitive with existing data hiding algorithms, and further that they can be made robust to noise: our models learn to reconstruct hidden information in an encoded image despite the presence of Gaussian blurring, pixel-wise dropout, cropping, and JPEG compression. Even though JPEG is non-differentiable, we show that a robust model can be trained using differentiable approximations. Finally, we demonstrate that adversarial training improves the visual quality of encoded images.
\keywords{Adversarial Networks, Steganography, Robust blind watermarking, Deep Learning, Convolutional Networks}
\end{abstract}

\section{Introduction}
Sometimes there is more to an image than meets the eye. An image may appear normal to a casual observer, but knowledgeable recipients can extract more information. Two common settings exist for hiding information in images. In \emph{steganography}, the goal is secret communication: a sender (Alice) encodes a message in an image such that the recipient (Bob) can decode the message, but an adversary (Eve) cannot tell whether any given image contains a message or not; Eve's task of detecting encoded images is called \emph{steganalysis}. In \emph{digital watermarking}, the goal is to encode information robustly: Alice wishes to encode a fingerprint in an image; Eve will then somehow distort the image (by cropping, blurring, etc), and Bob should be able to detect the fingerprint in the distorted image. Digital watermarking can be used to identify image ownership: if Alice is a photographer, then by embedding digital watermarks in her images she can prove ownership of those images even if versions posted online are modified.

Interestingly, neural networks are also capable of ``detecting'' information from images that are not visible to human eyes. Recent research have showed that neural networks are susceptible to \emph{adversarial examples}: given an image and a target class, the pixels of the image can be imperceptibly modified such that it is confidently classified as the target class~\cite{goodfellow2015explaining,szegedy2014intriguing}. Moreover, the adversarial nature of these generated images is preserved under a variety of image transformations~\cite{kurakin2016adversarial}. While the existence of adversarial examples is usually seen as a disadvantage of neural networks, it can be desirable for information hiding: if a network can be fooled with small perturbations into making incorrect class predictions, it should be possible to extract meaningful information from similar perturbations.

\begin{figure}[t]
\centering
\center
    \includegraphics[width=0.5\linewidth]{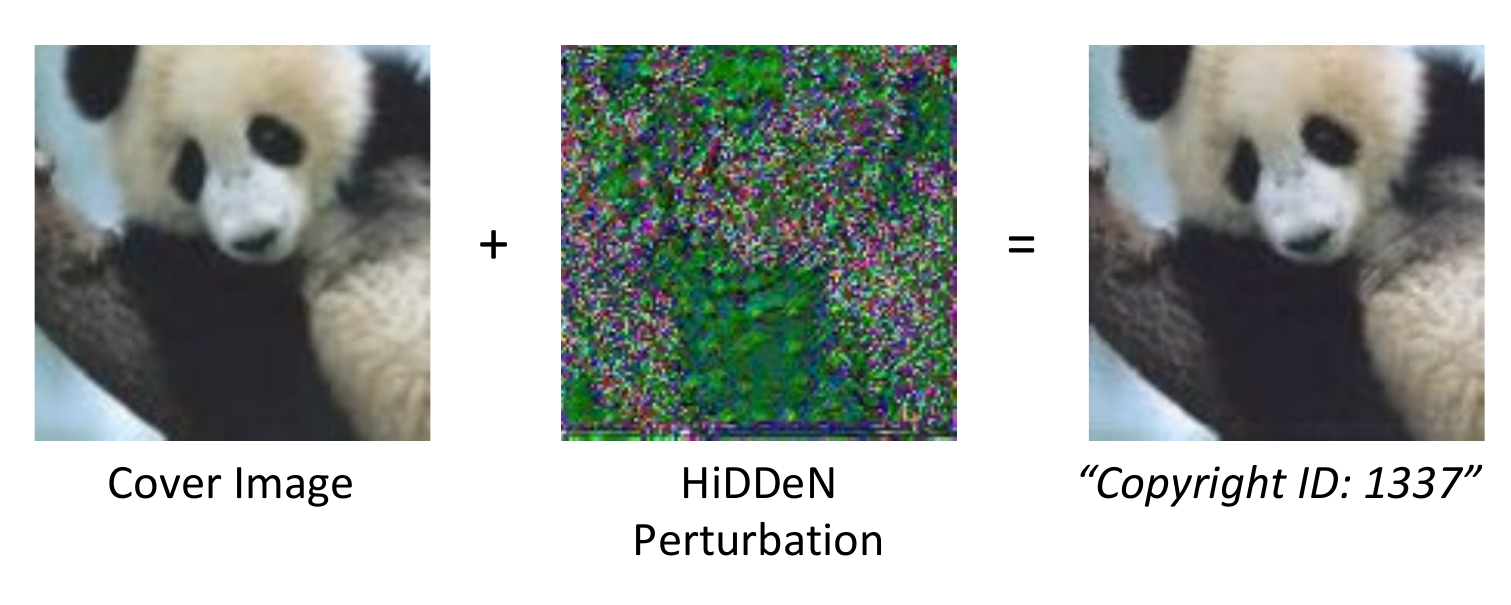}
   \vspace{-4mm}
  \caption{Given a cover image and a binary message, the HiDDeN encoder produces a visually indistinguishable \textit{encoded image} that contains the message, which can be recovered with high accuracy by the decoder.}
  \vspace{-5.5mm}
  \label{fig:pull}
\end{figure}

We introduce HiDDeN, the first end-to-end trainable framework for data hiding which can be applied to both steganography and watermarking. HiDDeN uses three convolutional networks for data hiding. An \emph{encoder} network receives a \emph{cover image} and a message (encoded as a bit string) and outputs an \emph{encoded image}; a \emph{decoder} network receives the encoded image and attempts to reconstruct the message. A third network, the \emph{adversary}, predicts whether a given image contains an encoded message; this provides an adversarial loss that improves the quality of encoded images. In many real world scenarios, images are distorted between a sender and recipient (e.g. during lossy compression). We model this by inserting optional \emph{noise layers} between the encoder and decoder, which apply different image transformations and force the model to learn encodings that can survive noisy transmission. We model the data hiding objective by minimizing (1) the difference between the cover and encoded images, (2) the difference between the input and decoded messages, and (3) the ability of an adversary to detect encoded images.
 
We analyze the performance of our method by measuring \emph{capacity}, the size of the message we can hide; \emph{secrecy}, the degree to which encoded images can be detected by steganalysis tools (\emph{steganalyzers}); and \emph{robustness}, how well our encoded messages can survive image distortions of various forms. We show that our methods outperform prior work in deep-learning-based steganography, and that our methods can also produce robust blind watermarks.
The networks learn to reconstruct hidden information in an encoded image despite the presence of Gaussian blurring, pixel-wise dropout, cropping, and JPEG compression. Though JPEG is not differentiable, we can reliably train networks that are robust to its perturbations using a differentiable approximation at training time.

Classical data hiding methods typically use heuristics to decide how much to modify each pixel. For example, some algorithms manipulate the least significant bits of some selected pixels~\cite{hugo}; others change mid-frequency components in the frequency domain~\cite{bi2007robust}. These heuristics are effective in the domains for which they are designed, but they are fundamentally \emph{static}. In contrast, HiDDeN can easily adapt to new requirements, since we directly optimize for the objectives of interest. For watermarking, one can simply retrain the model to gain robustness against a new type of noise instead of inventing a new algorithm. End-to-end learning is also advantageous in steganography, where having a diverse class of embedding functions (the same architecture, trained with different random initializations, produces very different embedding strategies) can stymie an adversary's ability to detect a hidden message.

\section{Related Work}
\paragraph{Adversarial examples.} Adversarial examples were shown to disrupt classification accuracy of various networks with minimal perturbation to the original images~\cite{szegedy2014intriguing}. They are typically computed by adding a small perturbation to each pixel in the direction that maximizes one output neuron~\cite{goodfellow2015explaining}. Adversarial examples generated for one network can transfer to another network~\cite{papernot2017practical}, suggesting that they come from a universal property of commonly used networks. Kurakin \etal~showed that adversarial examples are robust against image transformations; when an adversarial example is printed and photographed, the network still misclassifies the photo~\cite{kurakin2016adversarial}. Instead of injecting perturbations that lead to misclassification, we consider the possibility of transmitting useful information through adding the appropriate perturbations.

\paragraph{Steganography.}
A wide variety of steganography settings and methods have been proposed in the literature; most
relevant to our work are methods for \emph{blind image steganography}, where the message is encoded
in an image and the decoder does not have access to the original cover image. Least-Significant Bit (LSB) methods modify the lowest-order bits of each image
pixel depending on the bits of the secret message; several examples of LSB schemes are described in
\cite{van1994digital,wolfgang1996watermark}. By design, LSB methods produce image perturbations
which are not visually apparent. However, they can systematically alter the statistics of
the image, leading to reliable detection~\cite{Qin2010}.

Many steganography algorithms differ only in how they define a particular distortion metric to minimize during encoding. Highly Undetectable Steganography (HUGO)
\cite{hugo} measures distortion by computing weights for local pixel neighborhoods, resulting in lower distortion costs along edges and in high-texture regions.
WOW (Wavelet Obtained Weights)~\cite{wow} penalizes distortion to predictable regions of the image using a bank of directional filters.
S-UNIWARD~\cite{suniward} is similar to WOW but can be used for embedding in an arbitrary domain. 

\paragraph{Watermarking.}
Watermarking is similar to steganography: both aim to encode a secret message into an image. However, while the goal of steganography is secret communication, watermarking is frequently used to prove image ownership as a form of copyright protection. As such, watermarking methods prioritize robustness over secrecy: messages should be recoverable even after the encoded image is modified or distorted. \emph{Non-blind} methods assumes access to the unmodified cover image~\cite{cox1997secure,hsieh2001hiding,pereira2000robust}; more relevant to us are \emph{blind} methods~\cite{bi2007robust} where the decoder does not assume access to the cover image. Some watermarking methods encode information in the least significant bits of image pixels~\cite{van1994digital}; however for more robust encoding many methods instead encode information in the frequency domain~\cite{bi2007robust,hsieh2001hiding,pereira2000robust,potdar2005survey}. Other methods combine frequency-domain encoding with log-polar mapping~\cite{zheng2003rst} or template matching~\cite{pereira2000robust} to achieve robustness against spatial domain transformations.

\begin{figure}[t]
\centering
\includegraphics[width=0.9\textwidth]{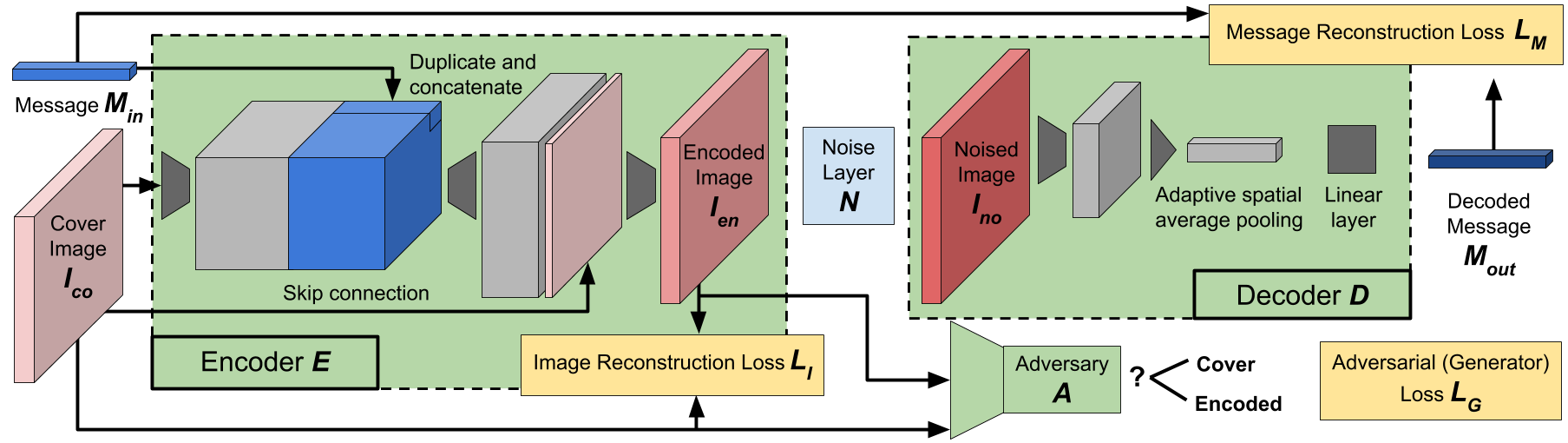}
\caption{
Model overview. The encoder $E$ receives the secret message $M$ and cover image $I_{co}$ as input and produces an encoded image $I_{en}$. The noise layer $N$ distorts the encoded image, producing a noised image $I_{no}$. The decoder produces a predicted message from the noised image. The adversary is trained to detect if an image is encoded.  The encoder and decoder are jointly trained to minimize loss $\mathcal{L}_I$ from difference between the cover and encoded image, loss $\mathcal{L}_M$ from difference between the input and predicted message and loss $\mathcal{L}_G$ from encoded image $I_{en}$ being detected by the adversary.
}
\vspace{-4mm}
\label{figure:system}
\end{figure}

\paragraph{Data Hiding with Neural Networks.}
\label{section:neuralDataHiding}
Neural networks have been used for both steganography and watermarking~\cite{isac2011study}. Until recently, prior work has typically used them for one stage of a larger
pipeline, such as determining watermarking strength per image region~\cite{jin2007applications},
or as part of the encoder~\cite{kandi2017exploring} or the decoder~\cite{mun2017robust}.

In contrast, we model the entire data hiding pipeline with neural networks and train them end-to-end. Different from~\cite{jin2007applications}, HiDDeN is a blind method: it does not require the recipient to have access to the original image, which is more useful than non-blind methods in many practical scenarios. \cite{mun2017robust} uses gradient descent to do encoding, whereas HiDDeN hides information in a single forward pass. \cite{hayes2017ste} is a recent end-to-end approach to steganography using adversarial networks, with which we compare results in Section~\ref{section:stego}. \cite{baluja2017hiding} train networks to hide an entire image within another image. In contrast to our work, neither of \cite{hayes2017ste,baluja2017hiding} consider encoding robustness; their focus is on steganography whereas HiDDeN can be used for watermarking as well.
 
Neural networks have also been applied to other forms of data hiding. Abadi and Anderson~\cite{abadi2016learning} show that adversarial networks can be trained for cryptography; 
Uchida \etal~\cite{uchida2017embedding} embed watermarks into the weights of trained neural networks; 
Fang \etal~\cite{tina2017generating} use LSTMs to embed secret messages into text.

\section{Method}
\label{section:Method}
Our goal is to develop a learnable, end-to-end model for image steganography and watermarking that can be made robust to arbitrary types of image distortion. To this end, our system comprises four main components: an \emph{encoder} $E_\theta$, a parameterless \emph{noise layer} $N$, a \emph{decoder} $D_\phi$ and an \emph{adversarial discriminator} $A_\gamma$. $\theta$, $\phi$ and $\gamma$ are trainable parameters. 
The encoder $E_\theta$ receives a \emph{cover image} $I_{co}$ of shape $C\times H\times W$ and a binary secret message $M_{in}\in\{0, 1\}^L$ of length $L$ and produces an \emph{encoded image} (also called the \emph{stego image}) $I_{en}$ of the same shape as $I_{co}$. 
The noise layer $N$ receives $I_{co}$ and $I_{en}$ as input and distorts the encoded image to produce a noised image $I_{no}$. 
The decoder $D$ recovers a message $M_{out}$ from $I_{no}$. 
At the same time, given an image $\tilde{I} \in \{I_{co}, I_{en}\}$, i.e.\ either a cover image or encoded image, the discriminator predicts $A(\tilde{I}) \in [0,1]$, the probability that $\tilde{I}$ is an encoded image. 

Encoded images should look visually similar to cover images. We characterize the ``similarity'' with \emph{image distortion loss}, the $l_2$ distance between $I_{co}$ and $I_{en}$: $\mathcal{L}_I(I_{co}, I_{en}) = \|I_{co} - I_{en}\|_2^2 / (CHW)$, 
and \emph{adversarial loss}, the ability of the discriminator to detect an encoded image $I_{en}$: $\mathcal{L}_{G}(I_{en}) = \log(1 - A(I_{en}))$~\cite{goodfellow2014generative}.
The discriminator incurs a classification loss from its predictions $\mathcal{L}_A(I_{co}, I_{en}) = \log(1 - A(I_{co})) + \log(A(I_{en}))$. 
The decoded message should be the same as the encoded message. We impose a \emph{message distortion loss} using the $l_2$ distance between the original and decoded messages $\mathcal{L}_M(M_{in}, M_{out}) = \|M_{in} - M_{out}\|_2^2/L$.
We perform stochastic gradient descent on $\theta, \phi$ to minimize the following loss over the distribution of input messages and images:
\begin{equation}
  \mathbb{E}_{I_{co}, M_{in}}\big[
  \mathcal{L}_M(M_{in}, M_{out}) + \lambda_I \mathcal{L}_I(I_{co}, I_{en})  + \lambda_G \mathcal{L}_{G}(I_{en})\big]
\end{equation}
where $\lambda_I$ and $\lambda_G$ control the relative weights of the losses. At the same time, we train discriminator $A_\gamma$ to minimize the following loss over the same distribution: 
\begin{equation}
  \mathbb{E}_{I_{co}, M_{in}}\left[\mathcal{L}_A(I_{co}, I_{en})\right].
\end{equation}

\paragraph{Network Architecture.} A diagram for our system setup is shown in Figure~\ref{figure:system}, and details can be found in Appendix A. The \textbf{encoder} first applies convolutions to input $I_{co}$ to form some intermediate representation. Next, we aim to incorporate the message input (of length $L$) in such a way that the encoder can easily learn to embed parts of it at any spatial location of the output. For this reason, we replicate the message spatially, and concatenate this ``message volume'' to the encoder's intermediary representation. This ensures that each convolutional filter in the next layer has access to the entire message as it convolves across each spatial location. After more convolutional layers, the encoder produces $I_{en}$, the encoded image. The \textbf{noise layer}, given $I_{co}, I_{en}$, applies noise and produces $I_{no}$. We do not require $I_{no}$ to have the same dimension as $I_{en}$, so that we can model dimension-altering noise like cropping. The \textbf{decoder} first applies several convolutional layers to produce $L$ feature channels in the intermediate representation. It then applies global spatial average pooling to produce a vector $L$ of the same size of the message -- the average pooling ensures that it can handle inputs of different spatial dimensions. It then applies a single linear layer to produce the predicted message. The \textbf{adversary} has a structure similar to the decoder, but it outputs a binary classification instead.

As we shall see, by varying message length $L$, image size $C, H, W$, noise layer $N$ and weight factors $\lambda_I$ and $\lambda_G$, the network setup described above can be adapted well to both steganography and watermarking. The specific choices will be discussed in their respective experiments.


%

\begin{figure*}[t]
\begin{center}
\begingroup
\setlength{\tabcolsep}{2pt} 
\small
\renewcommand{\arraystretch}{1} 
\scalebox{0.9}{
\begin{tabular}{ c  c  c  c  c  c  c c} 
 & Dropout & Cropout & Crop & Gaussian & real JPEG & JPEG & JPEG\\
 & $p=$30\% & $p=$30\% & $p=$3.5\% & $\sigma = 2$ & Q = 50 & Mask & Drop\\
$I_{en}$& 
\raisebox{-.5\height}{\includegraphics[width=0.12\textwidth]{NoiseExample/I1.png}} &
\raisebox{-.5\height}{\includegraphics[width=0.12\textwidth]{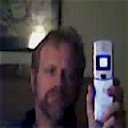}} &
\raisebox{-.5\height}{\includegraphics[width=0.12\textwidth]{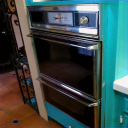}} &
\raisebox{-.5\height}{\includegraphics[width=0.12\textwidth]{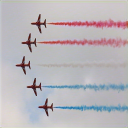}} &
\raisebox{-.5\height}{\includegraphics[width=0.12\textwidth]{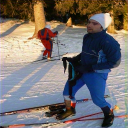}} &
\raisebox{-.5\height}{\includegraphics[width=0.12\textwidth]{NoiseExample/I6.png}} & 
\raisebox{-.5\height}{\includegraphics[width=0.12\textwidth]{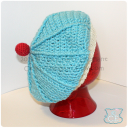}} 
\\
\noalign{\vskip 0.75mm}
$I_{no}$& 
\raisebox{-.5\height}{\includegraphics[width=0.12\textwidth]{NoiseExample/1.png}} &
\raisebox{-.5\height}{\includegraphics[width=0.12\textwidth]{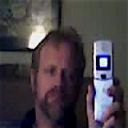}} &
\raisebox{-.5\height}{\includegraphics[width=0.12\textwidth]{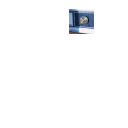}} &
\raisebox{-.5\height}{\includegraphics[width=0.12\textwidth]{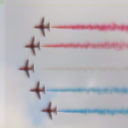}} &
\raisebox{-.5\height}{\includegraphics[width=0.12\textwidth]{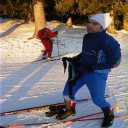}} &
\raisebox{-.5\height}{\includegraphics[width=0.12\textwidth]{NoiseExample/6.png}} &
\raisebox{-.5\height}{\includegraphics[width=0.12\textwidth]{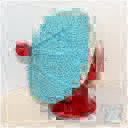}}
\\
\noalign{\vskip 0.75mm}
$15|I_{en} - I_{no}|$& 
\raisebox{-.5\height}{\includegraphics[width=0.12\textwidth]{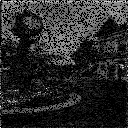}} &
\raisebox{-.5\height}{\includegraphics[width=0.12\textwidth]{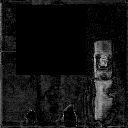}} &
\raisebox{-.5\height}{\includegraphics[width=0.12\textwidth]{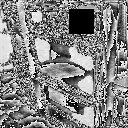}} &
\raisebox{-.5\height}{\includegraphics[width=0.12\textwidth]{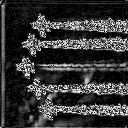}} &
\raisebox{-.5\height}{\includegraphics[width=0.12\textwidth]{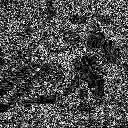}} &
\raisebox{-.5\height}{\includegraphics[width=0.12\textwidth]{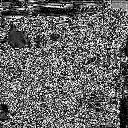}} &
\raisebox{-.5\height}{\includegraphics[width=0.12\textwidth]{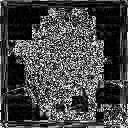}}
\end{tabular}
}
\endgroup
\vspace{1mm}
\caption{Illustration of non-identity noise layers. JPEG-Mask and JPEG-Drop are differentiable approximations of the JPEG compression (see Figure~\ref{figure:jpeg_simulation}). \textbf{Top:} Encoded image $I_{en}$. \textbf{Middle:} Noised image $I_{no}$. \textbf{Bottom:} Magnified difference $|I_{en}-I_{no}|$. 
Even under heavy distortion, such as a Crop layer which retains only 3.5\% of the original image, our model still learns to recover the watermark with high accuracy (see Section~\ref{section:Experiments}).
}
\vspace{-6mm}
\label{figure:transmitterExample}
\end{center}
\end{figure*}

\paragraph{Noise Layers.}  \label{section:noise}
\label{section:transmitterDefinition}
Our model should be able to hide images robustly against a variety of image distortions;
for this reason we consider six different types of noise layers which apply different types of
image distortion. Examples of all distortion types are shown in Figure~\ref{figure:transmitterExample}.

The \textbf{Identity} layer is the simplest: it leaves $I_{en}$ unchanged. 
The \textbf{Dropout} and \textbf{Cropout} layers undo some of the changes made by the encoder, producing the noisy image $I_{no}$ by combining pixels from the cover image $I_{co}$ and the encoded image $I_{en}$. Both types of noise keep a percentage $p$ of pixels from $I_{en}$ and use the rest from $I_{co}$, but Dropout makes this choice independently per pixel while Cropout keeps a random square crop from $I_{en}$. 
The \textbf{Gaussian} layer blurs the encoded image $I_{en}$ with a Gaussian kernel of width $\sigma$, and 
the \textbf{Crop} layer produces a random square $H' \times W'$ crop of the encoded image, where the ratio of image sizes $\frac{H' \times W'}{H \times W}$ is $p \in (0,1)$. 
The \textbf{JPEG} layer applies JPEG compression to $I_{en}$ with quality factor $Q \in (0, 100)$.  
Note that all non-identity noise layers have a scalar hyperparameter governing the \emph{intensity} of the distortion: Dropout, Cropout, and Crop keep a fraction $p$ of the pixels from $I_{en}$, Gaussian has a kernel width $\sigma$; and JPEG has a quality factor $Q$.

We experiment with an extensive family of distortions for the noise layer, $N$. Dropout, Crop and Cropout mainly affect the spatial domain, while Gaussian and JPEG act on the frequency domain. 
Granting the noise layer access to the cover image makes it more challenging as well. For an LSB algorithm, a noise layer that replaces each tampered pixel with a fixed value is analogous to a binary erasure channel, whereas a noise layer that replaces encoded pixels with original pixels acts as a binary symmetric channel. It is harder to be robust against the latter since the decoder has no information about where the tampering happens. 
Similarly, not only does the crop layer require the decoder to be input size agnostic, it also provides no information about where the $H' \times W'$ crop came from, further limiting the knowledge of the decoder.

\begin{figure}[t]
\centering
  \includegraphics[width=0.6\textwidth]{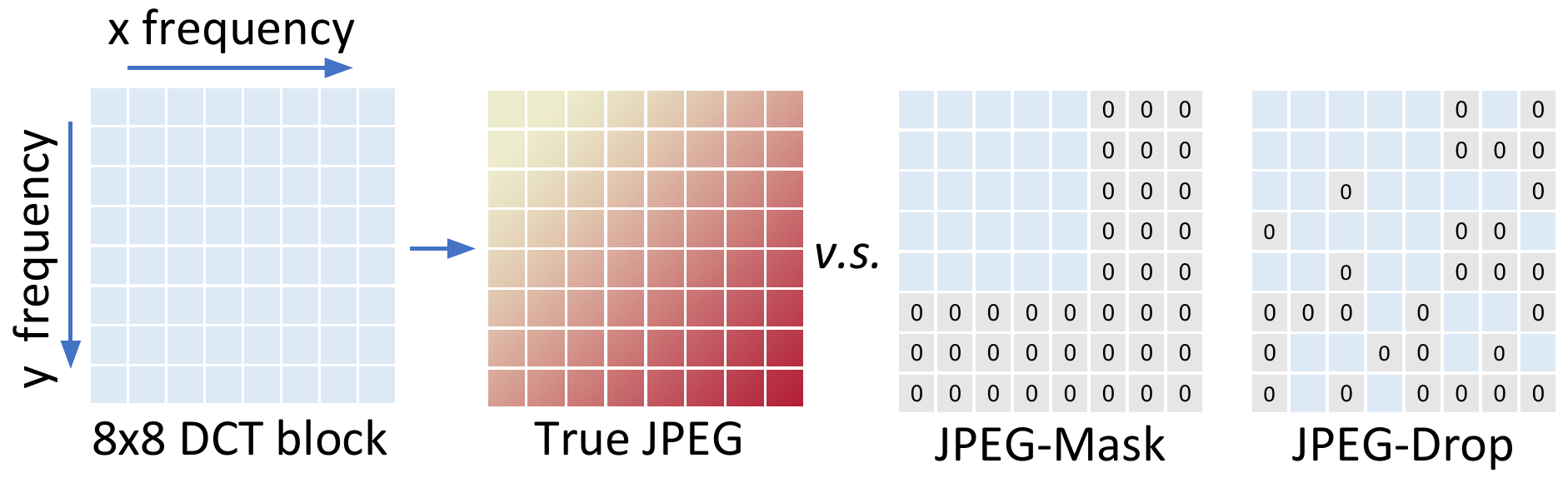}
  \caption{
    JPEG compresses an image by performing a discrete cosine transform (DCT) to give an $8\times8$ grid of frequency components (left) which are then quantized, with higher frequency components quantized more aggressively (middle left; bright red signifies stronger quantization). The DCT transform can be implemented as a single 8x8 stride 8 convolution layer with 64 filters fixed to the DCT basis, but due to the quantization step, JPEG compression is not differentiable. We therefore train models using two differentiable approximations: JPEG-Mask (middle right) zeros a fixed set of high-frequency coefficients, and JPEG-Drop (right) zeros channels with higher drop probabilities for high-frequency coefficients. Models trained against either approximation perform well against true JPEG compression at test time; see Figure~\ref{figure:jpeg_transfer}.
  }
  \vspace{-5mm}
  \label{figure:jpeg_simulation}
\end{figure}

\paragraph{``Differentiating'' the JPEG compression.} 
Although the network is trained with gradient descent, we do not need the test-time noise to be differentiable. We successfully obtain robustness against non-differentiable JPEG compression by training on differentiable approximations. JPEG compression divides the image into $8\times8$ regions, computes a discrete cosine transformation (DCT) within each region (essentially a change of basis in $\mathbb{R}^{64}$), then quantizes the resulting frequency-domain coefficients to different coarseness. Thus, only perceptually relevant information is preserved~\cite{wallace1992jpeg}. The quantization step is non-differentiable, rendering JPEG compression unfit for gradient-based optimization. 

Nevertheless, quantizing a value is information-theoretically equivalent to limiting the amount of information that can be transmitted though that ``channel''. To limit the amount of information being passed through specific frequency domain channels, we created noise layers that simulate the JPEG compression. These layers apply the DCT transformation using an $8\times8$ stride $8$ convolution layer with each filter corresponding to a basis vector in the DCT transformation. The network activations thus represent DCT domain coefficients of the encoded image. Masking/Dropout is then applied to the DCT coefficients to limit information flow; higher frequency coefficients are more likely to be masked/dropped, see Figure~\ref{figure:jpeg_simulation}. 
The noised image $I_{no}$ is then produced using a transpose convolution to implement the inverse DCT transform.

\begin{figure}[t]
\centering
\hfill \includegraphics[width = 0.37\textwidth]{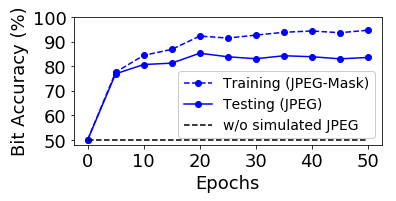}  \hfill \includegraphics[width = 0.37\textwidth]{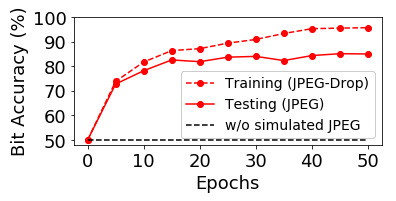} \hfill
\vspace{-4mm}
\caption{Bit accuracy for models trained with JPEG-Mask (blue, zero-masking on DCT coefficients) / JPEG-Drop (red, dropout on DCT coefficients). When trained against these approximations (dashed lines), both become robust against actual JPEG compression (solid lines, quality $Q = 50$).}
\vspace{-5mm}
\label{figure:jpeg_transfer}
\end{figure} 

We call the corresponding layers \textbf{JPEG-Mask} and \textbf{JPEG-Drop}. JPEG-Mask applies a fixed masking that only keeps 25 low frequency DCT coefficients in the Y channel and 9 in the U, V channels (following JPEG, which also preserves more information in the Y channel). The other coefficients are set to zero. JPEG-Drop applies a progressive dropout on the coefficients. The coarser the quantization for a coefficient in actual JPEG compression, the more likely this coefficient is zeroed in our simulation. Both methods successfully produce models that are robust against \emph{actual} JPEG compression, see Figure~\ref{figure:jpeg_transfer}.

\paragraph{Implementation details.}
All models are trained on 10,000 cover images from the COCO~\cite{lin2014microsoft} training set, resized to experiment-specific dimensions. Evaluation is performed on a 1000 image test set unseen during training. 
Messages are sampled with each bit drawn uniformly at random.
For gradient descent, we use Adam~\cite{kingma2014adam} with a learning rate of $10^{-3}$ and default hyperparameters. All models are trained with batch size 12. Models are trained for 200 epochs, or 400 epochs if being trained on multiple noise layers.

\section{Experiments}
\label{section:Experiments}
We evaluate our model on three axes: \emph{capacity}, the number of message bits which can be hidden per image bit; \emph{secrecy}, the difficulty of detecting our encoded images; and \emph{robustness}, the degree to which our method can succeed in the face of image distortions.

\paragraph{Metrics.} We use a number of metrics to measure our performance along these axes. We measure capacity primarily with \textbf{bits per pixel} (BPP), which is the number of message bits hidden per pixel of the encoded image; this is simply $L/(HWC)$ using the notation of Section~\ref{section:Method}. We measure secrecy primarily with \textbf{detection rate}; we train a modern steganalyzer~\cite{ats} which attempts to separate cover from stego images; its test-set accuracy is the detection rate. As a proxy to secrecy, we can also measure the image distortion between the cover and stego image using \textbf{peak signal-to-noise ratio} (PSNR); methods which distort the cover image more are likely to be more detectable. We measure robustness using \textbf{bit accuracy}, which is the number of identical bits between the input message $M$ and the predicted message $M'$, as we subject the model to varying image distortions between the encoder and decoder during testing.

Many of these metrics compete against each other. Models which have higher capacity typically sacrifice secrecy, since hiding more information in images naturally leads to larger image distortions; models that are very robust to noise typically sacrifice capacity or secrecy, since the message must be encoded in the image redundantly. In some sense, steganography and watermarking are at two ends of a spectrum of problems prioritizing these different axes; steganography stresses secrecy while watermarking emphasizes robustness. Since HiDDeN encompasses a whole family of data hiding algorithms, we can flexibly train models emphasizing different axes by the varying message length $L$, using different noise layers, and so on.

\subsection{Steganography: Capacity and Secrecy} \label{section:stego}

\begin{figure}[t]
\centering
  \scalebox{0.85}{
\begin{tabular}{c c c c c c} 
& Bits per & Bit & Detection rate \\
Method & pixel & error & (ATS~\cite{ats}) (\%) \\ [0.5ex] 
\hline \\ [-1.5ex]
HUGO~\cite{hugo} & 0.200 & -        &  70   \\
WOW~\cite{wow} & 0.200 & -        &  68   \\
S-UNIWARD~\cite{suniward} & 0.200 & -   &  68   \\ 
HiDDeN (model weights known) & 0.203 & $< 10^{-5}$ &   98   \\ 
HiDDeN (model weights unknown) & 0.203 & $< 10^{-5}$  &  \textbf{50}   \\ 
\hline 
\end{tabular}}
\\*[1mm]
\begingroup
\setlength{\tabcolsep}{1pt} 
\renewcommand{\arraystretch}{1} 
\begin{tabular}{cccccc}
\includegraphics[width=0.75in]{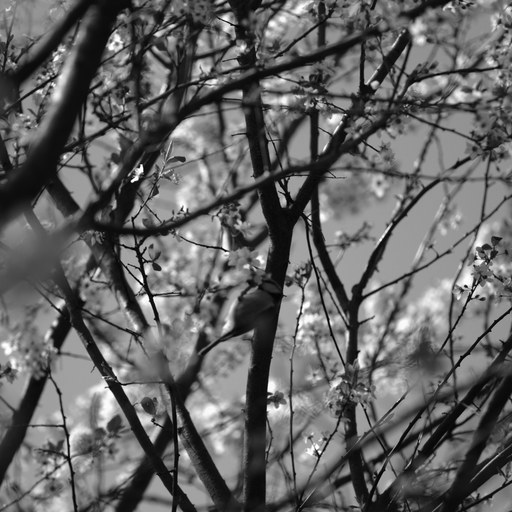} %
&
\includegraphics[width=0.75in]{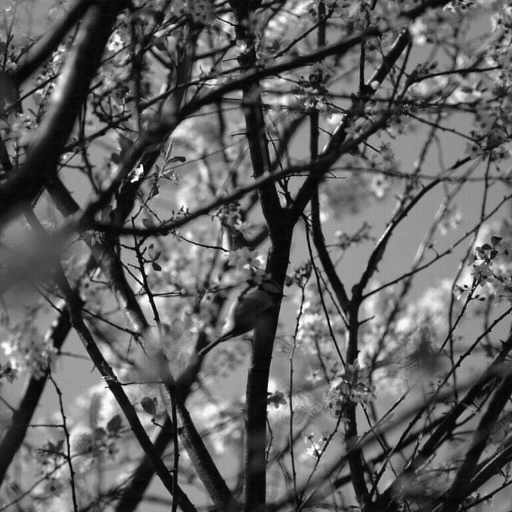} %
&
\includegraphics[width=0.75in]{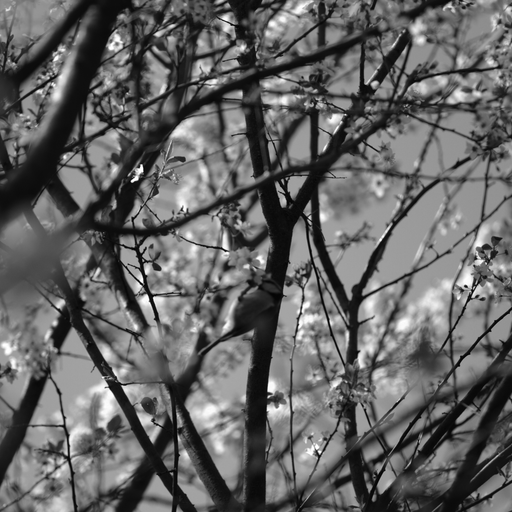} %
&
\includegraphics[width=0.75in]{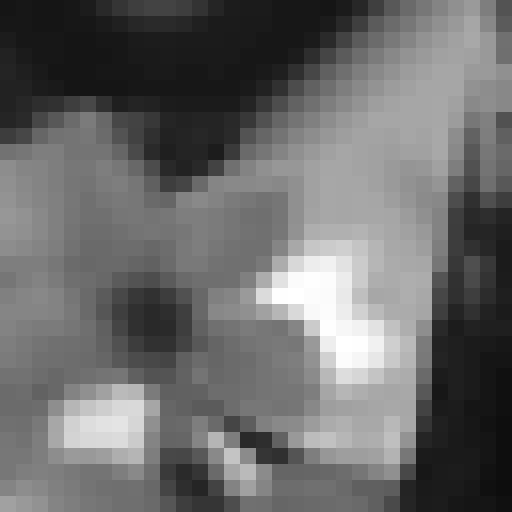}
&
\includegraphics[width=0.75in]{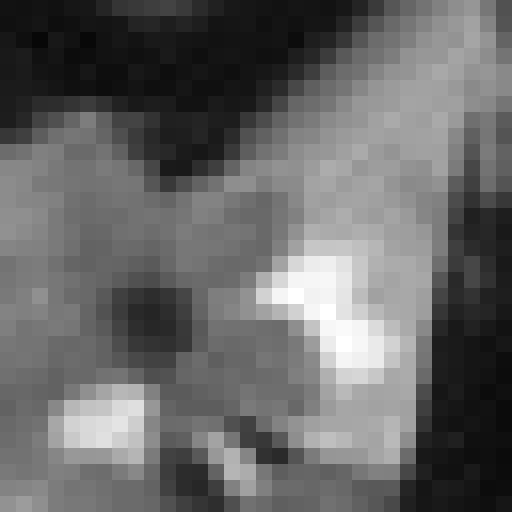}
&
\includegraphics[width=0.75in]{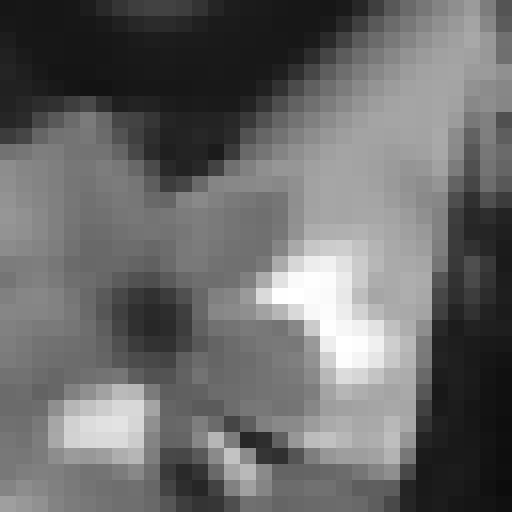}
\\
Cover & Our Method & HUGO & Cover & Our Method & HUGO\\
(a)  &  (b) & (c) & (d) &  (e) & (f) 
\end{tabular}
\endgroup
\vspace{1mm}
\caption{
\textbf{Top}: 
Capacity and secrecy of HiDDeN versus classical steganography methods. With comparable capacity to baselines as measured by bits per pixel and bit accuracy, our method is undetectable (50\%, no better than chance) by ATS under the assumption that an adversary has access to the algorithm but not the trained model weights.
\textbf{Bottom:}
Example of $512\times512$ encoded images. (a) Cover, (b) encoded by our method, (c) encoded by HUGO \cite{hugo}. (d), (e), (f) are $32\times32$ crops of (a), (b), (c).
}
\vspace{-4mm}
\label{figure:stegoGenerated}
\end{figure}

\begin{figure}
\centering
\begin{footnotesize}
\scalebox{0.9}{
\begin{tabular}{cccc}

\raisebox{3mm}{\shortstack{Cover \\ Image }} & 
\raisebox{3mm}{\shortstack{Ours \\ $0.1$ bpp }} & 
\raisebox{3mm}{\shortstack{Ours \\ $0.2$ bpp }} & 
\raisebox{3mm}{\shortstack{\cite{hayes2017ste} \\ $0.1$ bpp }} 
\vspace{-2mm}
\\ [-1ex]
\hline \\ [-2.5ex]
  Error: & $<10^{-5}$ & $< 10^{-5}$ & $2 \times 10^{-3}$
\\
\includegraphics[width=0.75in]{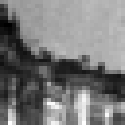} &
\includegraphics[width=0.75in]{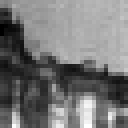} &
\includegraphics[width=0.75in]{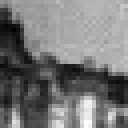} &
\includegraphics[width=0.75in]{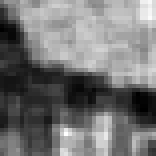} \\

\includegraphics[width=0.75in]{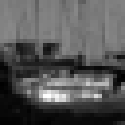} &
\includegraphics[width=0.75in]{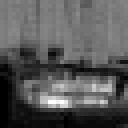} &
\includegraphics[width=0.75in]{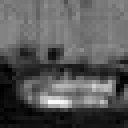} &
\includegraphics[width=0.75in]{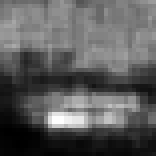} \\

\includegraphics[width=0.75in]{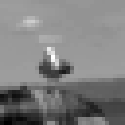} &
\includegraphics[width=0.75in]{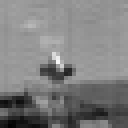} &
\includegraphics[width=0.75in]{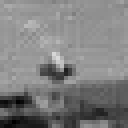} &
\includegraphics[width=0.75in]{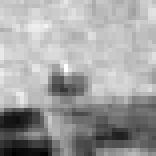} \\
  
\end{tabular}
}
\vspace{-2mm}
\end{footnotesize}
  \caption{Comparison of our encoding method against \cite{hayes2017ste}. We encode twice as many bits in images of the same size, while having smaller decoding error and better encoded image quality.}
\label{fig:compare}

\vspace{-4mm}
\end{figure}

We train our model to encode binary messages of length $L=52$ in grayscale images of size $16\times16$, giving our trained model a capacity of $52/(16\times16)\approx0.203$ BPP. For larger images, we use a longer message to preserve encoding capacity. Our model encodes an $L$-bit chunk of the message on each $16 \times 16$ patch of the image. Steganography typically assumes noiseless transmission, so we set $N$ to be the identity layer. We first compare with traditional methods \cite{hugo,wow,suniward} in Figure~\ref{figure:stegoGenerated} and then with a neural-network-based approach \cite{hayes2017ste} in Figure~\ref{fig:compare}.

\paragraph{Image quality} We compare with three recent steganography methods: HUGO~\cite{hugo}, WOW~\cite{wow}, and S-UNIWARD~\cite{suniward}. Each is configured with a capacity of 0.2 BPP to closely match our model. We test all methods on $512\times512$ cover images from the BOSS dataset~\cite{bas2011break}, which is a common testbed for steganography algorithms. Since our model was trained on COCO, this showcases its generality to new data. Figure~\ref{figure:stegoGenerated} shows example cover and stego images from our method and from HUGO (other classical baselines give results indistinguishable from HUGO). Both HUGO and our method generate encoded images which are visually indistinguishable from the cover image. 

\paragraph{Capacity.} Our model closely matches the baselines on BPP (0.203 vs 0.200). One disadvantage of neural networks in steganography is the possibility for the decoder network to make mistakes. As shown in Figure~\ref{figure:stegoGenerated}, our method reconstructs the message accurately but not perfectly, having an error rate of $<10^{-5}$. One can use error correcting codes if higher accuracy is desired.

\paragraph{Secrecy.} We measure secrecy by training the steganalyzer ATS~\cite{ats} to distinguish between cover and stego images for all methods; results are shown in Figure~\ref{figure:stegoGenerated}. Following the recommendation of the authors for best performance~\cite{ats}, we provide ATS with small datasets for each algorithm: 250 cover and 250 stego images. 

ATS generates an eponymous \emph{artificial training set} from this unlabeled test set by repeatedly applying the steganography algorithm to the data, then uses this artificial training set to train a supervised classifier. Finally, the trained classifier is evaluated on its ability to recognize stego images in the test set.

For our model we evaluate two threat scenarios. We first train one encoder-decoder pair and evaluate ATS assuming it has access to the learned model directly. In this setting (model weights known), ATS achieves a detection accuracy of 98\%, showing the susceptibility of our model to statistical analysis when the precise weights are known. In the second threat scenario, we consider the case where the attacker knows HiDDeN is being used (and furthermore, knows the precise architecture, hyperparameters and dataset) but does not have access to the trained model itself (which is not exactly reproducible without the random seed used to initialize the parameters). To test this scenario, we train five identical HiDDeN models with different seeds and provide these models to ATS for building its artificial training set. Finally, we train a sixth model with a new random initialization for evaluation. When we test ATS on this sixth model (Ours -- different model), it achieves a detection accuracy of 50\%, no better than random guessing. In comparison, ATS achieves detection accuracies of 70\% for HUGO and 68\% for WOW and S-UNIWARD. This result shows that image distortion patterns produced by each trained model are unique, which makes our approach undetectable by ATS without access to the trained model directly. This hints at an advantage of learned steganography models: unlike static classical algorithms, they can employ a diverse set of encoding strategies.

\setlength{\tabcolsep}{1pt} 

\paragraph{Against other neural network based methods} 
Compared to \cite{hayes2017ste} which uses a fully connected network to generate encoded images, our method uses convolutional networks, greatly improving encoded image quality. Figure~\ref{fig:compare} compares our results with \cite{hayes2017ste}; at double their bit rate we achieve lower error and generate images much closer to the cover image.

\subsection{Watermarking: Robustness}
\label{section:robustness}
Digital watermarking prioritizes robustness over capacity and secrecy; it hides only a small message in the image, but that information should remain even after significant distortions are applied to the encoded image. By varying the type of image distortion applied at training time, we show that our model can learn robustness to a variety of different image distortions.

\begin{figure*}
\begin{center}
\small
\scalebox{0.9}{\begin{tabular}{c c c c c c c c c c} 
 & Digimarc & Identity & Dropout & Cropout & Crop & Gaussian & JPEG-mask & JPEG-drop & Combined \\ [0.5ex] 
\hline \\ [-1.5ex]
PSNR(Y) & 62.12 & 44.63 & 42.52 & 47.24 & 35.20 & 40.55 & 30.09 & 28.79 & 33.55\\
PSNR(U) & 38.23 & 45.44 & 38.52 & 40.97 & 33.31 & 41.96 & 35.33 & 32.51 & 38.92\\
PSNR(V) & 52.06 & 46.90 & 41.05 & 41.88 & 35.86 & 42.88 & 36.27 & 33.42 & 39.38\\
[1ex] 
\hline 
\end{tabular}}
\\*[1mm]
\begingroup
\setlength{\tabcolsep}{2pt} 
\renewcommand{\arraystretch}{1} 
\begin{tabular}{cc | ccc | c}
\multicolumn{2}{c|}{} & \multicolumn{3}{|c|}{\scalebox{0.9}{Trained with Adversary}} & \multicolumn{1}{|c}{\scalebox{0.9}{No Adversary}}\\
\hline
Cover & Digimarc & Crop & Gaussian & Combined & Combined\\
\includegraphics[width=0.15\textwidth]{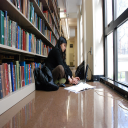} & 
\includegraphics[width=0.15\textwidth]{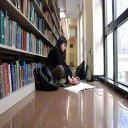} & 
\includegraphics[width=0.15\textwidth]{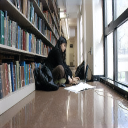} & 
\includegraphics[width=0.15\textwidth]{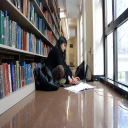} & 
 \includegraphics[width=0.15\textwidth]{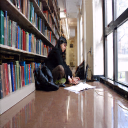} & 
 \includegraphics[width=0.15\textwidth]{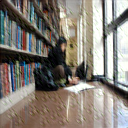} \\ 
\end{tabular}
\endgroup

\caption{
Image distortions for watermarking algorithms. \textbf{Top:} Mean PSNR between cover and encoded images for Digimarc and our model trained with different noise layers. \textbf{Bottom:} A cover image and encoded images from both Digimarc and our model trained with Crop, Gaussian, and Combined noise layers. \textbf{Bottom Right:} An encoded image from a model trained with combined noise but without an adversary. Adversarial training significantly improves the visual quality of the encoded images. 
}
\vspace{-5mm}
\label{figure:imageDistortion}

\end{center}
\end{figure*}

We train our model on YUV color images of size $C \times H \times W = 3 \times 128 \times 128$ with message length $L=30$ and weight factors $\lambda_I = 0.7$ and $\lambda_G = 0.001$. A model is trained on the identity noise layer as reference. Six models are trained using different noise layers $N$ at training time: Dropout ($p=0.3$), Cropout ($p=0.3$), Crop ($p=0.035$), Gaussian ($\sigma=2.0$), JPEG-mask and JPEG-drop. We call these types of noise-resistant models \emph{specialized models} as they are trained to be robust against a particular kind of noise. We additionally train a noise-resistant \emph{combined} model by using a different noise layer for each mini-batch during training. See the supplementary materials for details.

\paragraph{Baseline.} 
To the best of our knowledge, there are no open source implementations of recent methods for digital watermarking. As a baseline we compare to Digimarc~\cite{digimarc}, a closed source commercial package for digital watermarking. Since Digimarc is closed source, there are certain limitations when comparing HiDDeN against it, especially for comparing transmission accuracy. Detailed analysis and comparison methodology are provided in the appendix.

\paragraph{Qualitative Results.}
Figure~\ref{figure:imageDistortion} shows qualitative examples of $128\times128$ images encoded with each of our trained models, as well as a $128\times128$ image encoded with Digimarc. For each image we report the PSNR between the cover image $I_{co}$ and the encoded image $I_{en}$. We see that encoded images from our models are visually indistinguishable from the cover image, and that we can train a single model (Combined) that is simultaneously robust to all types of noise without sacrificing image quality.

\paragraph{Adversary.}
Figure~\ref{figure:imageDistortion} also compares generated images of two models, one trained with the adversary and the other trained without the adversary. Both models are trained on the combined noise layer and tuned individually. 
The model trained with $l_2$ loss alone has visible artifacts, as shown in the rightmost image of Figure~\ref{figure:imageDistortion}. The model trained against an adversarial discriminator produces images with no visible artifacts (Figure~\ref{figure:imageDistortion}, second image from the right).

\begin{figure}[t]
\centering
\hspace*{-1.5mm}\includegraphics[width=0.85\textwidth]{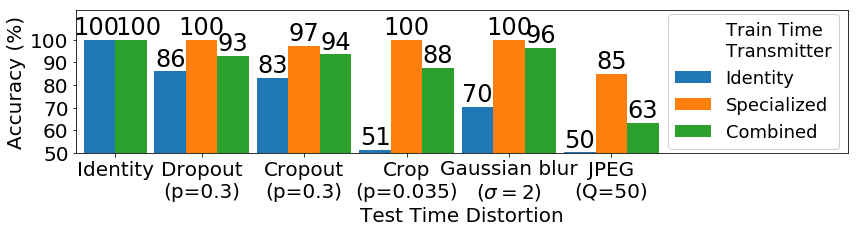}
\vspace{-3mm}
\caption{
Robustness of our models against different test time distortions. Each cluster uses a different test time distortion. Identity (blue) is trained with no image distortion; Specialized (orange) is trained on the same type of distortion used during testing; Combined (green) is trained on all types of distortions.
}
\label{figure:robustness}
\end{figure}

\begin{figure}[t]
\centering
\includegraphics[width = 0.9\textwidth]{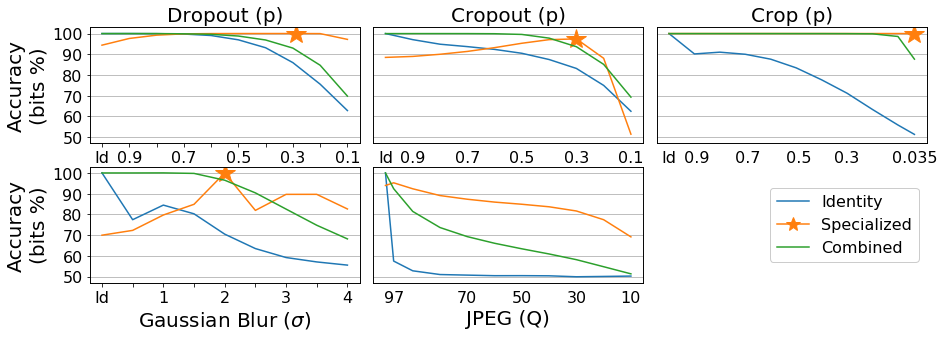}
\vspace{-3mm}
\caption{Bit accuracy under various distortions and intensities. Stars denote the noise intensity used during training. The specialized JPEG model is trained on the differentiable approximation JPEG-Mask, and the plot shows performance on actual JPEG.
}
\vspace{-2mm}
\label{figure:intensity_dep}
\end{figure}

\paragraph{Robustness.}
The intensity of an image distortion can be controlled with a scalar: keep percentage $p$ for Dropout, Cropout, and Crop, kernel width $\sigma$ for Gaussian, and quality $Q$ for JPEG compression. Figure~\ref{figure:robustness} shows the bit accuracy of models when they are tested on various noise layers. For each tested noise layer, we evaluate the model trained with the identity noise layer, i.e. no noise (blue), the model trained on the same noise layer (orange), and the model trained on combined noise layers (green). Bit accuracies are measured on 1000 images unseen during training. Figure~\ref{figure:intensity_dep} reports bit accuracy as a function of test time distortion intensity.

The model trained without noise unsurprisingly performs poorly when tested against different noise layers, and fails completely ($50\%$ bit accuracy, no better than chance) when tested on Crop and JPEG. Since this model enjoyed lossless transmission between the encoder and decoder during training, it has no incentive to learn robustness to any type of image distortion.

However, the high bit accuracies of the Specialized models (orange bars) in Figure~\ref{figure:robustness} demonstrate that models \textit{can} learn robustness to many different types of image distortion when these distortions are introduced into the training process. 
This remains true even when the distortion is non-differentiable:
Models trained without noise have $50\%$ bit accuracy when tested against true JPEG compression, but this improves to $85\%$ when trained with simulated JPEG noise.

Finally, we see that in most cases the Combined model, which is trained on all types of noise, is competitive with specialized models despite its increased generality. For example, it achieves $94\%$ accuracy against Cropout, close to the $97\%$ accuracy of the specialized model.

\begin{figure}[t]
\centering
\includegraphics[width = 0.85\textwidth]{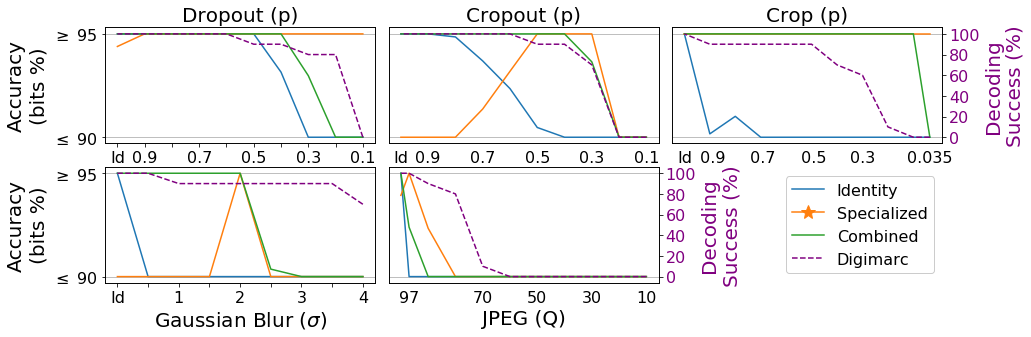}
\vspace{-3mm}
\caption{Model performance under different distortions and intensities. 
We compare the model trained with no noise (blue), models specialized to a particular distortion (orange), and a Combined model trained on all distortion types (green). We also show Digimarc's decoding success rate for $256\times256$ images (purple). The two axis are scaled to translate bit accuracy into full reconstruction rate. See Appendix B for detail.
}
\vspace{-3mm}
\label{figure:digimarc_comp}
\end{figure}
 

\paragraph{Comparison with Digimarc.} Digimarc is closed source, and it only reports success or failure for decoding a fixed-size watermark. It provides no information about its bit error rate, which makes comparing it with HiDDeN difficult. 

To ensure a fair comparison, we first estimate the capacity of Digimarc, and then apply an error correcting code that matches HiDDeN's bit rate with Digimarc. This also allows us to converts bit accuracy to decode success rate since a few errors can be corrected (see Appendix B for full methodology). 
From this analysis, we consider $\geq 95\%$ bit accuracy for our model to be comparable to a successful Digimarc decoding, and $\leq 90\%$ bit accuracy to be a failed decoding.

We report the comparison in Figure~\ref{figure:digimarc_comp}, with the y-axis clipped according to our analysis. For spatial domain noise, our model exceeds the performance of Digimarc at high noise intensities. Against Dropout ($p=0.1$), our specialized model has bit accuracy $\geq 95\%$, yet Digimarc fails completely. Against Crop ($p=0.1$), both the specialized and combined models have bit accuracy $\geq 95\%$, but Digimarc cannot reconstruct any of the ten watermarks tested. For frequency domain noise, our model performs worse than Digimarc. This is likely due to the fact that we baked no assumptions about frequency domain transformations into the architecture, whereas watermarking tools commonly work directly in the frequency domain.

\section{Conclusion}
We have developed a framework for data hiding in images which is trained end-to-end using neural networks. Compared to classical data hiding methods, ours allows flexibly trading off between capacity, secrecy, and robustness to different types of noise by varying parameters or noise layers at training-time. Compared to deep learning methods for steganography, we demonstrate improved quantitative and qualitative performance. For robust watermarking, HiDDeN is to our knowledge the first end-to-end method using neural networks. Ultimately, end-to-end methods like HiDDeN have a fundamental advantage in robust  data-hiding: new distortions can be incorporated directly into the training procedure, with no need to design new, specialized algorithms. In future work, we hope to see improvements in message capacity, robustness to more diverse types of image distortions -- such as geometric transforms, contrast change, and other lossy compression schemes -- and procedures for data hiding in other input domains, such as audio and video.

\textbf{Acknowledgments} Our work is supported by an ONR MURI grant. We would like to thank Ehsan Adeli, Rishi Bedi, Jim Fan, Kuan Fang, Adithya Ganesh, Agrim Gupta, De-An Huang, Ranjay Krishna, Damian Mrowca, Ben Zhang and anonymous reviewers for their feedback on our work.

\clearpage

\bibliographystyle{splncs}
\bibliography{main}

\begin{thebibliography}{10}

\bibitem{goodfellow2015explaining}
Goodfellow, I.J., Shlens, J., Szegedy, C.:
\newblock Explaining and harnessing adversarial examples.
\newblock In: ICLR. (2015)

\bibitem{szegedy2014intriguing}
Szegedy, C., Zaremba, W., Sutskever, I., Bruna, J., Erhan, D., Goodfellow, I.,
  Fergus, R.:
\newblock Intriguing properties of neural networks.
\newblock In: ICLR. (2014)

\bibitem{kurakin2016adversarial}
Kurakin, A., Goodfellow, I., Bengio, S.:
\newblock Adversarial examples in the physical world.
\newblock arXiv preprint arXiv:1607.02533 (2016)

\bibitem{hugo}
Pevn{\'y}, T., Filler, T., Bas, P.
\newblock In: Using High-Dimensional Image Models to Perform Highly
  Undetectable Steganography. Springer Berlin Heidelberg, Berlin, Heidelberg
  (2010)  161--177

\bibitem{bi2007robust}
Bi, N., Sun, Q., Huang, D., Yang, Z., Huang, J.:
\newblock Robust image watermarking based on multiband wavelets and empirical
  mode decomposition.
\newblock IEEE Transactions on Image Processing (2007)

\bibitem{papernot2017practical}
Papernot, N., McDaniel, P., Goodfellow, I., Jha, S., Celik, Z.B., Swami, A.:
\newblock Practical black-box attacks against machine learning.
\newblock In: Proceedings of the 2017 ACM on Asia Conference on Computer and
  Communications Security, ACM (2017)  506--519

\bibitem{van1994digital}
Van~Schyndel, R.G., Tirkel, A.Z., Osborne, C.F.:
\newblock A digital watermark.
\newblock In: IEEE Converence on Image Processing, 1994, IEEE (1994)

\bibitem{wolfgang1996watermark}
Wolfgang, R.B., Delp, E.J.:
\newblock A watermark for digital images.
\newblock In: Image Processing, 1996. Proceedings., International Conference
  on. Volume~3., IEEE (1996)  219--222

\bibitem{Qin2010}
Qin, J., Xiang, X., Wang, M.X.:
\newblock A review on detection of {LSB} matching steganography.
\newblock Information Technology Journal \textbf{9}(8) (aug 2010)  1725--1738

\bibitem{wow}
Holub, V., Fridrich, J.:
\newblock Designing steganographic distortion using directional filters.
\newblock In: 2012 IEEE International Workshop on Information Forensics and
  Security (WIFS). (Dec 2012)  234--239

\bibitem{suniward}
Holub, V., Fridrich, J., Denemark, T.:
\newblock Universal distortion function for steganography in an arbitrary
  domain.
\newblock EURASIP Journal on Information Security \textbf{2014}(1) (2014) ~1

\bibitem{cox1997secure}
Cox, I.J., Kilian, J., Leighton, F.T., Shamoon, T.:
\newblock Secure spread spectrum watermarking for multimedia.
\newblock IEEE transactions on image processing \textbf{6}(12) (1997)
  1673--1687

\bibitem{hsieh2001hiding}
Hsieh, M.S., Tseng, D.C., Huang, Y.H.:
\newblock Hiding digital watermarks using multiresolution wavelet transform.
\newblock IEEE Transactions on industrial electronics \textbf{48}(5) (2001)
  875--882

\bibitem{pereira2000robust}
Pereira, S., Pun, T.:
\newblock Robust template matching for affine resistant image watermarks.
\newblock IEEE transactions on image Processing \textbf{9}(6) (2000)
  1123--1129

\bibitem{potdar2005survey}
Potdar, V., Han, S., Chang, E.:
\newblock A survey of digital image watermarking techniques.
\newblock In: 3rd IEEE International Conference on Industrial Informatics
  (INDIN 2005), IEEE (2005)  709--716

\bibitem{zheng2003rst}
Zheng, D., Zhao, J., El~Saddik, A.:
\newblock Rst-invariant digital image watermarking based on log-polar mapping
  and phase correlation.
\newblock IEEE transactions on circuits and systems for video technology (2003)

\bibitem{isac2011study}
Isac, B., Santhi, V.:
\newblock A study on digital image and video watermarking schemes using neural
  networks.
\newblock International Journal of Computer Applications \textbf{12}(9) (2011)
  1--6

\bibitem{jin2007applications}
Jin, C., Wang, S.:
\newblock Applications of a neural network to estimate watermark embedding
  strength.
\newblock In: Workshop on Image Analysis for Multimedia Interactive Services,
  IEEE (2007)

\bibitem{kandi2017exploring}
Kandi, H., Mishra, D., Gorthi, S.R.S.:
\newblock Exploring the learning capabilities of convolutional neural networks
  for robust image watermarking.
\newblock Computers \& Security (2017)

\bibitem{mun2017robust}
Mun, S.M., Nam, S.H., Jang, H.U., Kim, D., Lee, H.K.:
\newblock A robust blind watermarking using convolutional neural network.
\newblock arXiv preprint arXiv:1704.03248 (2017)

\bibitem{hayes2017ste}
Hayes, J., Danezis, G.:
\newblock ste-gan-ography: Generating steganographic images via adversarial
  training.
\newblock arXiv preprint arXiv:1703.00371 (2017)

\bibitem{baluja2017hiding}
Baluja, S.:
\newblock Hiding images in plain sight: Deep steganography.
\newblock In: Advances in Neural Information Processing Systems. (2017)
  2066--2076

\bibitem{abadi2016learning}
Abadi, M., Andersen, D.G.:
\newblock Learning to protect communications with adversarial neural
  cryptography.
\newblock arXiv preprint arXiv:1610.06918 (2016)

\bibitem{uchida2017embedding}
Uchida, Y., Nagai, Y., Sakazawa, S., Satoh, S.:
\newblock Embedding watermarks into deep neural networks.
\newblock arXiv preprint arXiv:1701.04082 (2017)

\bibitem{tina2017generating}
Fang, T., Jaggi, M., Argyraki, K.:
\newblock Generating steganographic text with lstms.
\newblock In: ACL Student Research Workshop 2017. Number EPFL-CONF-229881
  (2017)

\bibitem{goodfellow2014generative}
Goodfellow, I., Pouget-Abadie, J., Mirza, M., Xu, B., Warde-Farley, D., Ozair,
  S., Courville, A., Bengio, Y.:
\newblock Generative adversarial nets.
\newblock In: Advances in Neural Information Processing Systems. (2014)
  2672--2680

\bibitem{wallace1992jpeg}
Wallace, G.K.:
\newblock The jpeg still picture compression standard.
\newblock IEEE transactions on consumer electronics \textbf{38}(1) (1992)
  xviii--xxxiv

\bibitem{lin2014microsoft}
Lin, T.Y., Maire, M., Belongie, S., Hays, J., Perona, P., Ramanan, D.,
  Doll{\'a}r, P., Zitnick, C.L.:
\newblock Microsoft coco: Common objects in context.
\newblock In: ECCV. (2014)

\bibitem{kingma2014adam}
Kingma, D., Ba, J.:
\newblock Adam: A method for stochastic optimization.
\newblock In: ICLR. (2015)

\bibitem{ats}
Lerch-Hostalot, D., Megías, D.:
\newblock Unsupervised steganalysis based on artificial training sets.
\newblock Engineering Applications of Artificial Intelligence \textbf{50}
  (2016)  45 -- 59

\bibitem{bas2011break}
Bas, P., Filler, T., Pevn{\`y}, T.:
\newblock ” break our steganographic system”: The ins and outs of
  organizing boss.
\newblock In: International Workshop on Information Hiding, Springer (2011)
  59--70

\bibitem{digimarc}
Digimarc:
\newblock Digimarc.
\newblock \url{https://www.digimarc.com/home}

\bibitem{radfordnealLDPC}
Neal, R.M.:
\newblock Software for low density parity check codes.
\newblock \url{https://github.com/radfordneal/LDPC-codes} (2012)

\end{thebibliography}


\newpage

\appendix

\begin{figure}[t]
\centering
{\huge
HiDDeN: Hiding Data \\ with Deep Networks \\ Supplementary Material
}
\end{figure}

\section{Model Architecture}\label{section:architecture}

We denote the combination of Convolution, Batch Normalization and ReLU as a Conv-BN-ReLU block. In our experiments, all Conv-BN-ReLU blocks, unless otherwise specified, have $3 \times 3$ kernels, stride 1, and padding 1. 

Given input $I_{co}$ of shape $C \times H \times W$, the \textbf{encoder} applies four Conv-BN-ReLU blocks with 64 output filters to the input image, resulting in a $64 \times H \times W$ image activation volume. It then replicates $M$ spatially $H \times W$ times to form a $L \times H\times W$ message volume. The two volumes, along with the original image, are concatenated channel-wise into a single $(64 + L + C) \times H \times W$ activation. The encoder then applies a Conv-BN-ReLU block with $64$ output filters. A final convolution layer  with a $1 \times 1$ kernel, stride 1, no padding and $C$ output filters is used to produce $I_{en}$ with shape $C \times H \times W$. No activation function is applied after the final convolution. 

The \textbf{noise layer}, given $I_{co}, I_{en}$, produces $I_{no}$ with shape $C \times H' \times W'$. The decoder is required to support cases where $H' \neq H, W' \neq W$, e.g. when the image is cropped. 

The \textbf{decoder} contains 7 Conv-BN-ReLU blocks with 64 filters each and one last Conv-BN-ReLU block with $L$ filters. An average is pooling performed over all spatial dimensions and a final ($L \times L$) linear layer produces the predicted message $M_{out}$. Because of the use of average pooling, the decoder assumes nothing about $H'$ and $W'$. To evaluate the accuracy of $M_{out}$ during testing, we round each entry to 0 or 1 before comparing it with $M_{in}$. 

The \textbf{adversary} has a structure similar to the decoder but has fewer convolutions. It contains 3 Conv-BN-ReLU blocks with 64 filters each. The activation volume is averaged over spatial dimensions and a linear layer with two output units produces the logits for the two-class classification problem. 

\section{Digimarc Baseline}
\label{section:digimarcDetail}
To the best of our knowledge, there are no open source implementations of recent methods for digital watermarking. As a baseline we compare to Digimarc~\cite{digimarc}, a closed source commercial package for digital watermarking. 

Since Digimarc is closed source, we do not know the exact size of the message it encodes. A Digimarc watermark must encode a $6$ digit ID, $2$ digit PIN, $32$ bit integer, its corresponding class (one out of three choices), and three booleans. It also needs to identify that the decoded data is uncorrupted. We estimate that the these requirements can be met with a $64$ bit encoding. 

As users, we do not have access to the input and decoded messages from Digimarc; so we cannot report accuracies of each bit. Instead the Digimarc decoder simply reports ``success'' or ``failure'' for each image. So we only know if the full message is correctly reconstructed. As a coarse proxy we measure the fraction of images where Digimarc successfully decodes the message. Digimarc must be manually applied to each image, so results are averaged over 10 images.

Note that in our training our model encodes a $30$ bit message in a $128 \times 128$ image. However, Digimarc recommends users to encode on images at least of size $256 \times 256$. On these images, Digimarc's watermark only has $\frac{1}{2}$ of our bit rate. As a result, compared to Digimarc ($256$), one could apply error correction codes with capacity $\frac{1}{2}$ to further reduce error. Thus we consider using a LDPC-$36$-$128$ code \cite{radfordnealLDPC} with $\frac{1}{2}$ capacity to match the bit rate of the two methods. Preliminary experiments show that the error correction code can reconsturct the full code-word correctly when channel bit error rate is $\leq 5\%$, the ability to fully reconstruct the message drops to $< 10\%$ when the channel bit error rate is $\geq 10\%$. So we plot our models' bit accuracy against Digimarc's decoding success rate where $\geq 95\%$ accuracy corresponds to $100\%$ perfect reconstruction and $\leq 90\%$ accuracy corresponds to $0\%$ perfect reconstruction. Results are shown in Figure~\ref{figure:digimarc_comp}.

\section{Combined Model Training Details}
The ``Combined" model discussed in the second experiment is trained with a collection of noise layers that are swapped in each minibatch. It is a natural generalization from a model that learns to be robust against one specific noise to a model that learns robustness to multiple noises at the same time. We use various noise layers during training of the Combined model: Dropout $p = 0.3$, Dropout $p = 0.7$, Cropout $p = 0.3$, Cropout $p = 0.7$, Crop $p = 0.3$, Crop $p = 0.7$, Gaussian blur $\sigma = 2$, Gaussian blur $\sigma = 4$, JPEG-drop, JPEG-mask and the Identity layer.

\newpage

\section{More examples of encoded images}
\label{section:robustness:moreImages}
\begin{center}
\begingroup
\setlength{\tabcolsep}{1pt} 
\renewcommand{\arraystretch}{1} 
\begin{figure*}
\tiny
\begin{tabular}{ccccccccc}
Cover & Identity & Dropout & Cropout & Crop & Gaussian & JPEG-mask & JPEG-drop & Combined\\

\includegraphics[width=0.11\textwidth]{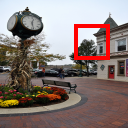} & 
 \includegraphics[width=0.11\textwidth]{generated_imgs_robustness/in_01_identity.png} & 
 \includegraphics[width=0.11\textwidth]{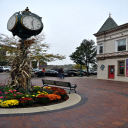} & 
 \includegraphics[width=0.11\textwidth]{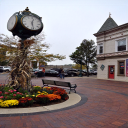} & 
 \includegraphics[width=0.11\textwidth]{generated_imgs_robustness/in_01_crop.png} & 
 \includegraphics[width=0.11\textwidth]{generated_imgs_robustness/in_01_gaussian.png} & 
 \includegraphics[width=0.11\textwidth]{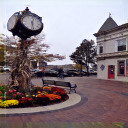} & 
 \includegraphics[width=0.11\textwidth]{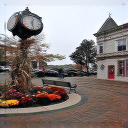} & 
 \includegraphics[width=0.11\textwidth]{generated_imgs_robustness/in_01_combined.png}\\

\includegraphics[width=0.11\textwidth]{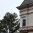} & 
 \includegraphics[width=0.11\textwidth]{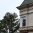} & 
 \includegraphics[width=0.11\textwidth]{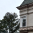} & 
 \includegraphics[width=0.11\textwidth]{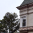} & 
 \includegraphics[width=0.11\textwidth]{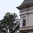} & 
 \includegraphics[width=0.11\textwidth]{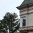} & 
 \includegraphics[width=0.11\textwidth]{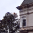} & 
 \includegraphics[width=0.11\textwidth]{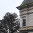} & 
 \includegraphics[width=0.11\textwidth]{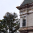}\\


\includegraphics[width=0.11\textwidth]{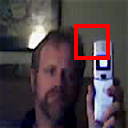} & 
 \includegraphics[width=0.11\textwidth]{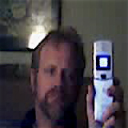} & 
 \includegraphics[width=0.11\textwidth]{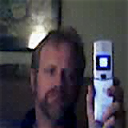} & 
 \includegraphics[width=0.11\textwidth]{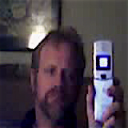} & 
 \includegraphics[width=0.11\textwidth]{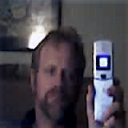} & 
 \includegraphics[width=0.11\textwidth]{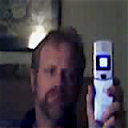} & 
 \includegraphics[width=0.11\textwidth]{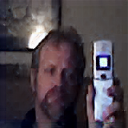} & 
 \includegraphics[width=0.11\textwidth]{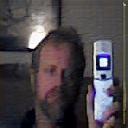} & 
 \includegraphics[width=0.11\textwidth]{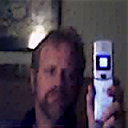}\\

\includegraphics[width=0.11\textwidth]{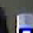} & 
 \includegraphics[width=0.11\textwidth]{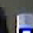} & 
 \includegraphics[width=0.11\textwidth]{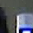} & 
 \includegraphics[width=0.11\textwidth]{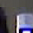} & 
 \includegraphics[width=0.11\textwidth]{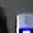} & 
 \includegraphics[width=0.11\textwidth]{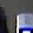} & 
 \includegraphics[width=0.11\textwidth]{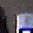} & 
 \includegraphics[width=0.11\textwidth]{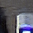} & 
 \includegraphics[width=0.11\textwidth]{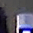}\\


\includegraphics[width=0.11\textwidth]{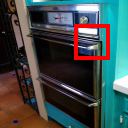} & 
 \includegraphics[width=0.11\textwidth]{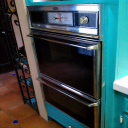} & 
 \includegraphics[width=0.11\textwidth]{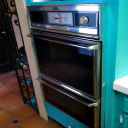} & 
 \includegraphics[width=0.11\textwidth]{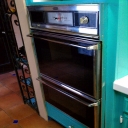} & 
 \includegraphics[width=0.11\textwidth]{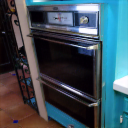} & 
 \includegraphics[width=0.11\textwidth]{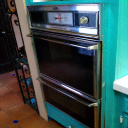} & 
 \includegraphics[width=0.11\textwidth]{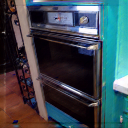} & 
 \includegraphics[width=0.11\textwidth]{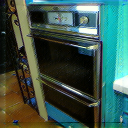} & 
 \includegraphics[width=0.11\textwidth]{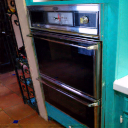}\\

\includegraphics[width=0.11\textwidth]{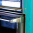} & 
 \includegraphics[width=0.11\textwidth]{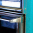} & 
 \includegraphics[width=0.11\textwidth]{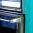} & 
 \includegraphics[width=0.11\textwidth]{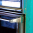} & 
 \includegraphics[width=0.11\textwidth]{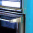} & 
 \includegraphics[width=0.11\textwidth]{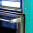} & 
 \includegraphics[width=0.11\textwidth]{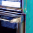} & 
 \includegraphics[width=0.11\textwidth]{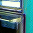} & 
 \includegraphics[width=0.11\textwidth]{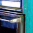}\\


\includegraphics[width=0.11\textwidth]{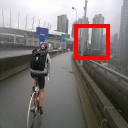} & 
 \includegraphics[width=0.11\textwidth]{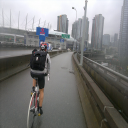} & 
 \includegraphics[width=0.11\textwidth]{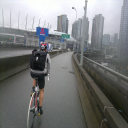} & 
 \includegraphics[width=0.11\textwidth]{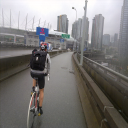} & 
 \includegraphics[width=0.11\textwidth]{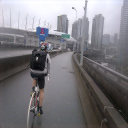} & 
 \includegraphics[width=0.11\textwidth]{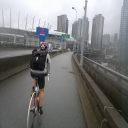} & 
 \includegraphics[width=0.11\textwidth]{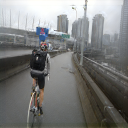} & 
 \includegraphics[width=0.11\textwidth]{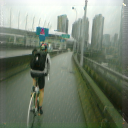} & 
 \includegraphics[width=0.11\textwidth]{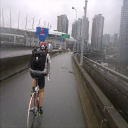}\\

\includegraphics[width=0.11\textwidth]{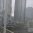} & 
 \includegraphics[width=0.11\textwidth]{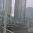} & 
 \includegraphics[width=0.11\textwidth]{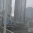} & 
 \includegraphics[width=0.11\textwidth]{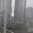} & 
 \includegraphics[width=0.11\textwidth]{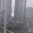} & 
 \includegraphics[width=0.11\textwidth]{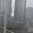} & 
 \includegraphics[width=0.11\textwidth]{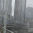} & 
 \includegraphics[width=0.11\textwidth]{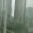} & 
 \includegraphics[width=0.11\textwidth]{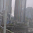}\\


\includegraphics[width=0.11\textwidth]{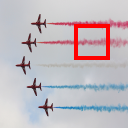} & 
 \includegraphics[width=0.11\textwidth]{generated_imgs_robustness/in_05_identity.png} & 
 \includegraphics[width=0.11\textwidth]{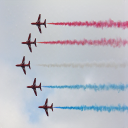} & 
 \includegraphics[width=0.11\textwidth]{generated_imgs_robustness/in_05_crop-out.png} & 
 \includegraphics[width=0.11\textwidth]{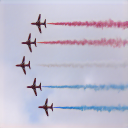} & 
 \includegraphics[width=0.11\textwidth]{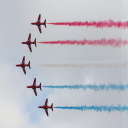} & 
 \includegraphics[width=0.11\textwidth]{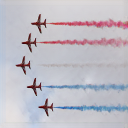} & 
 \includegraphics[width=0.11\textwidth]{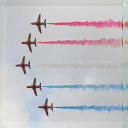} & 
 \includegraphics[width=0.11\textwidth]{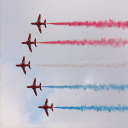}\\

\includegraphics[width=0.11\textwidth]{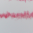} & 
 \includegraphics[width=0.11\textwidth]{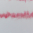} & 
 \includegraphics[width=0.11\textwidth]{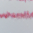} & 
 \includegraphics[width=0.11\textwidth]{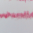} & 
 \includegraphics[width=0.11\textwidth]{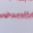} & 
 \includegraphics[width=0.11\textwidth]{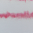} & 
 \includegraphics[width=0.11\textwidth]{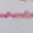} & 
 \includegraphics[width=0.11\textwidth]{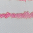} & 
 \includegraphics[width=0.11\textwidth]{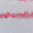}\\


\end{tabular}
\end{figure*}

\begin{figure*}
\tiny
\begin{tabular}{ccccccccc}
Cover & Identity & Dropout & Cropout & Crop & Gaussian & JPEG-mask & JPEG-drop & Combined\\

\includegraphics[width=0.11\textwidth]{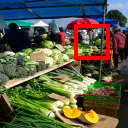} & 
 \includegraphics[width=0.11\textwidth]{generated_imgs_robustness/in_06_identity.png} & 
 \includegraphics[width=0.11\textwidth]{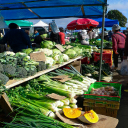} & 
 \includegraphics[width=0.11\textwidth]{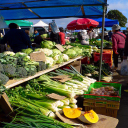} & 
 \includegraphics[width=0.11\textwidth]{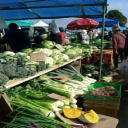} & 
 \includegraphics[width=0.11\textwidth]{generated_imgs_robustness/in_06_gaussian.png} & 
 \includegraphics[width=0.11\textwidth]{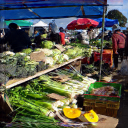} & 
 \includegraphics[width=0.11\textwidth]{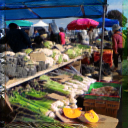} & 
 \includegraphics[width=0.11\textwidth]{generated_imgs_robustness/in_06_combined.png}\\

\includegraphics[width=0.11\textwidth]{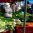} & 
 \includegraphics[width=0.11\textwidth]{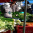} & 
 \includegraphics[width=0.11\textwidth]{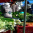} & 
 \includegraphics[width=0.11\textwidth]{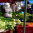} & 
 \includegraphics[width=0.11\textwidth]{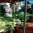} & 
 \includegraphics[width=0.11\textwidth]{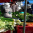} & 
 \includegraphics[width=0.11\textwidth]{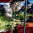} & 
 \includegraphics[width=0.11\textwidth]{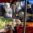} & 
 \includegraphics[width=0.11\textwidth]{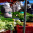}\\


\includegraphics[width=0.11\textwidth]{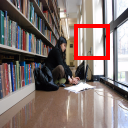} & 
 \includegraphics[width=0.11\textwidth]{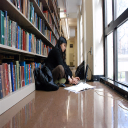} & 
 \includegraphics[width=0.11\textwidth]{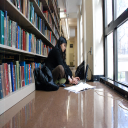} & 
 \includegraphics[width=0.11\textwidth]{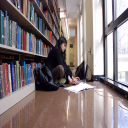} & 
 \includegraphics[width=0.11\textwidth]{generated_imgs_robustness/in_10_crop.png} & 
 \includegraphics[width=0.11\textwidth]{generated_imgs_robustness/in_10_gaussian.png} & 
 \includegraphics[width=0.11\textwidth]{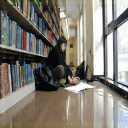} & 
 \includegraphics[width=0.11\textwidth]{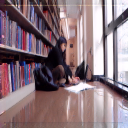} & 
 \includegraphics[width=0.11\textwidth]{generated_imgs_robustness/in_10_combined.png}\\

\includegraphics[width=0.11\textwidth]{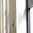} & 
 \includegraphics[width=0.11\textwidth]{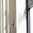} & 
 \includegraphics[width=0.11\textwidth]{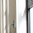} & 
 \includegraphics[width=0.11\textwidth]{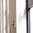} & 
 \includegraphics[width=0.11\textwidth]{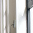} & 
 \includegraphics[width=0.11\textwidth]{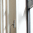} & 
 \includegraphics[width=0.11\textwidth]{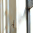} & 
 \includegraphics[width=0.11\textwidth]{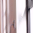} & 
 \includegraphics[width=0.11\textwidth]{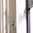}\\


\includegraphics[width=0.11\textwidth]{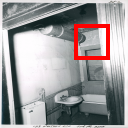} & 
 \includegraphics[width=0.11\textwidth]{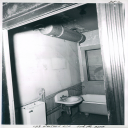} & 
 \includegraphics[width=0.11\textwidth]{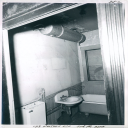} & 
 \includegraphics[width=0.11\textwidth]{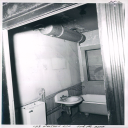} & 
 \includegraphics[width=0.11\textwidth]{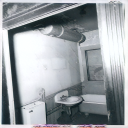} & 
 \includegraphics[width=0.11\textwidth]{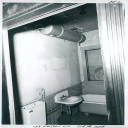} & 
 \includegraphics[width=0.11\textwidth]{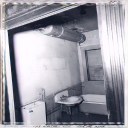} & 
 \includegraphics[width=0.11\textwidth]{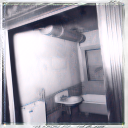} & 
 \includegraphics[width=0.11\textwidth]{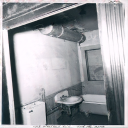}\\

\includegraphics[width=0.11\textwidth]{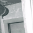} & 
 \includegraphics[width=0.11\textwidth]{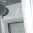} & 
 \includegraphics[width=0.11\textwidth]{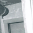} & 
 \includegraphics[width=0.11\textwidth]{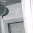} & 
 \includegraphics[width=0.11\textwidth]{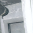} & 
 \includegraphics[width=0.11\textwidth]{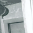} & 
 \includegraphics[width=0.11\textwidth]{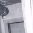} & 
 \includegraphics[width=0.11\textwidth]{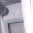} & 
 \includegraphics[width=0.11\textwidth]{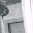}\\


\includegraphics[width=0.11\textwidth]{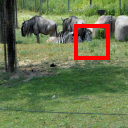} & 
 \includegraphics[width=0.11\textwidth]{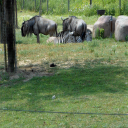} & 
 \includegraphics[width=0.11\textwidth]{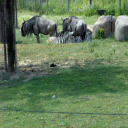} & 
 \includegraphics[width=0.11\textwidth]{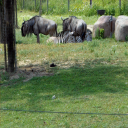} & 
 \includegraphics[width=0.11\textwidth]{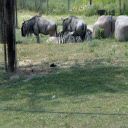} & 
 \includegraphics[width=0.11\textwidth]{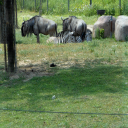} & 
 \includegraphics[width=0.11\textwidth]{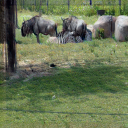} & 
 \includegraphics[width=0.11\textwidth]{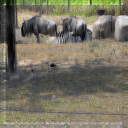} & 
 \includegraphics[width=0.11\textwidth]{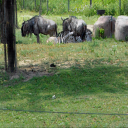}\\

\includegraphics[width=0.11\textwidth]{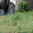} & 
 \includegraphics[width=0.11\textwidth]{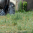} & 
 \includegraphics[width=0.11\textwidth]{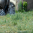} & 
 \includegraphics[width=0.11\textwidth]{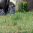} & 
 \includegraphics[width=0.11\textwidth]{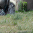} & 
 \includegraphics[width=0.11\textwidth]{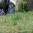} & 
 \includegraphics[width=0.11\textwidth]{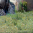} & 
 \includegraphics[width=0.11\textwidth]{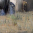} & 
 \includegraphics[width=0.11\textwidth]{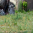}\\


\includegraphics[width=0.11\textwidth]{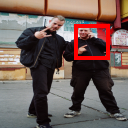} & 
 \includegraphics[width=0.11\textwidth]{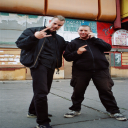} & 
 \includegraphics[width=0.11\textwidth]{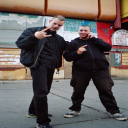} & 
 \includegraphics[width=0.11\textwidth]{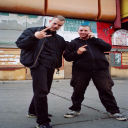} & 
 \includegraphics[width=0.11\textwidth]{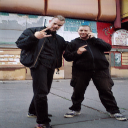} & 
 \includegraphics[width=0.11\textwidth]{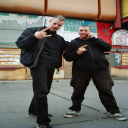} & 
 \includegraphics[width=0.11\textwidth]{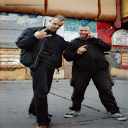} & 
 \includegraphics[width=0.11\textwidth]{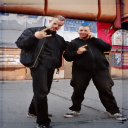} & 
 \includegraphics[width=0.11\textwidth]{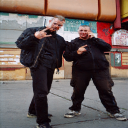}\\

\includegraphics[width=0.11\textwidth]{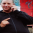} & 
 \includegraphics[width=0.11\textwidth]{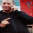} & 
 \includegraphics[width=0.11\textwidth]{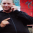} & 
 \includegraphics[width=0.11\textwidth]{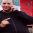} & 
 \includegraphics[width=0.11\textwidth]{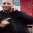} & 
 \includegraphics[width=0.11\textwidth]{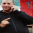} & 
 \includegraphics[width=0.11\textwidth]{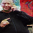} & 
 \includegraphics[width=0.11\textwidth]{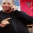} & 
 \includegraphics[width=0.11\textwidth]{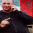}\\


\end{tabular}
\end{figure*}

\begin{figure*}
\tiny
\begin{tabular}{ccccccccc}
Cover & Identity & Dropout & Crop-out & Crop & Gaussian & JPEG-mask & JPEG-drop & Combined\\

\includegraphics[width=0.11\textwidth]{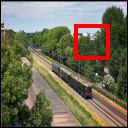} & 
 \includegraphics[width=0.11\textwidth]{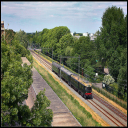} & 
 \includegraphics[width=0.11\textwidth]{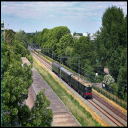} & 
 \includegraphics[width=0.11\textwidth]{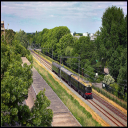} & 
 \includegraphics[width=0.11\textwidth]{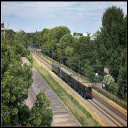} & 
 \includegraphics[width=0.11\textwidth]{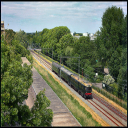} & 
 \includegraphics[width=0.11\textwidth]{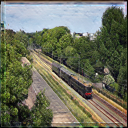} & 
 \includegraphics[width=0.11\textwidth]{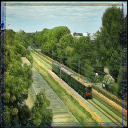} & 
 \includegraphics[width=0.11\textwidth]{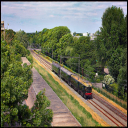}\\

\includegraphics[width=0.11\textwidth]{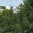} & 
 \includegraphics[width=0.11\textwidth]{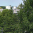} & 
 \includegraphics[width=0.11\textwidth]{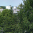} & 
 \includegraphics[width=0.11\textwidth]{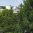} & 
 \includegraphics[width=0.11\textwidth]{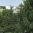} & 
 \includegraphics[width=0.11\textwidth]{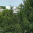} & 
 \includegraphics[width=0.11\textwidth]{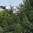} & 
 \includegraphics[width=0.11\textwidth]{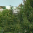} & 
 \includegraphics[width=0.11\textwidth]{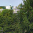}\\


\includegraphics[width=0.11\textwidth]{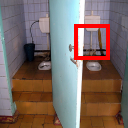} & 
 \includegraphics[width=0.11\textwidth]{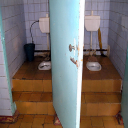} & 
 \includegraphics[width=0.11\textwidth]{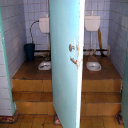} & 
 \includegraphics[width=0.11\textwidth]{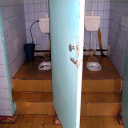} & 
 \includegraphics[width=0.11\textwidth]{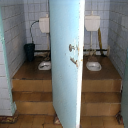} & 
 \includegraphics[width=0.11\textwidth]{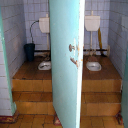} & 
 \includegraphics[width=0.11\textwidth]{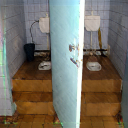} & 
 \includegraphics[width=0.11\textwidth]{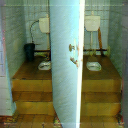} & 
 \includegraphics[width=0.11\textwidth]{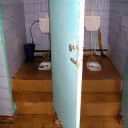}\\

\includegraphics[width=0.11\textwidth]{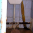} & 
 \includegraphics[width=0.11\textwidth]{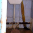} & 
 \includegraphics[width=0.11\textwidth]{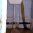} & 
 \includegraphics[width=0.11\textwidth]{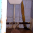} & 
 \includegraphics[width=0.11\textwidth]{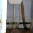} & 
 \includegraphics[width=0.11\textwidth]{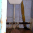} & 
 \includegraphics[width=0.11\textwidth]{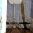} & 
 \includegraphics[width=0.11\textwidth]{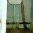} & 
 \includegraphics[width=0.11\textwidth]{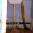}\\


\includegraphics[width=0.11\textwidth]{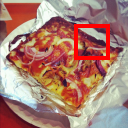} & 
 \includegraphics[width=0.11\textwidth]{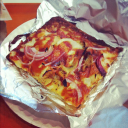} & 
 \includegraphics[width=0.11\textwidth]{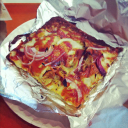} & 
 \includegraphics[width=0.11\textwidth]{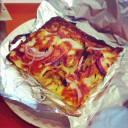} & 
 \includegraphics[width=0.11\textwidth]{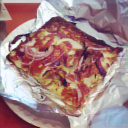} & 
 \includegraphics[width=0.11\textwidth]{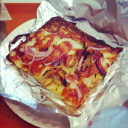} & 
 \includegraphics[width=0.11\textwidth]{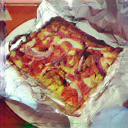} & 
 \includegraphics[width=0.11\textwidth]{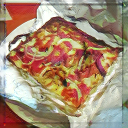} & 
 \includegraphics[width=0.11\textwidth]{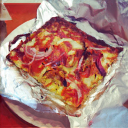}\\

\includegraphics[width=0.11\textwidth]{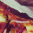} & 
 \includegraphics[width=0.11\textwidth]{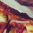} & 
 \includegraphics[width=0.11\textwidth]{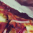} & 
 \includegraphics[width=0.11\textwidth]{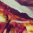} & 
 \includegraphics[width=0.11\textwidth]{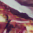} & 
 \includegraphics[width=0.11\textwidth]{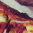} & 
 \includegraphics[width=0.11\textwidth]{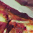} & 
 \includegraphics[width=0.11\textwidth]{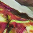} & 
 \includegraphics[width=0.11\textwidth]{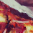}\\


\includegraphics[width=0.11\textwidth]{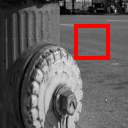} & 
 \includegraphics[width=0.11\textwidth]{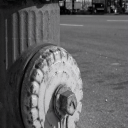} & 
 \includegraphics[width=0.11\textwidth]{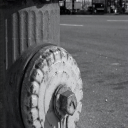} & 
 \includegraphics[width=0.11\textwidth]{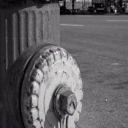} & 
 \includegraphics[width=0.11\textwidth]{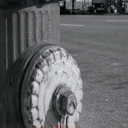} & 
 \includegraphics[width=0.11\textwidth]{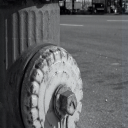} & 
 \includegraphics[width=0.11\textwidth]{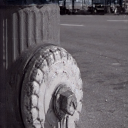} & 
 \includegraphics[width=0.11\textwidth]{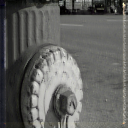} & 
 \includegraphics[width=0.11\textwidth]{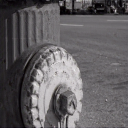}\\

\includegraphics[width=0.11\textwidth]{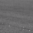} & 
 \includegraphics[width=0.11\textwidth]{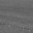} & 
 \includegraphics[width=0.11\textwidth]{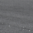} & 
 \includegraphics[width=0.11\textwidth]{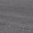} & 
 \includegraphics[width=0.11\textwidth]{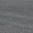} & 
 \includegraphics[width=0.11\textwidth]{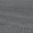} & 
 \includegraphics[width=0.11\textwidth]{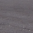} & 
 \includegraphics[width=0.11\textwidth]{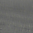} & 
 \includegraphics[width=0.11\textwidth]{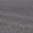}\\


\includegraphics[width=0.11\textwidth]{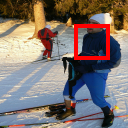} & 
 \includegraphics[width=0.11\textwidth]{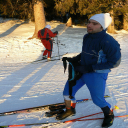} & 
 \includegraphics[width=0.11\textwidth]{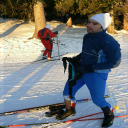} & 
 \includegraphics[width=0.11\textwidth]{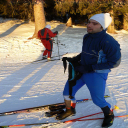} & 
 \includegraphics[width=0.11\textwidth]{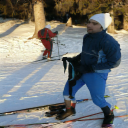} & 
 \includegraphics[width=0.11\textwidth]{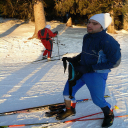} & 
 \includegraphics[width=0.11\textwidth]{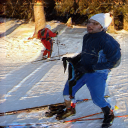} & 
 \includegraphics[width=0.11\textwidth]{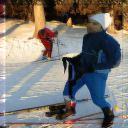} & 
 \includegraphics[width=0.11\textwidth]{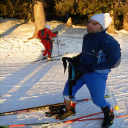}\\

\includegraphics[width=0.11\textwidth]{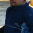} & 
 \includegraphics[width=0.11\textwidth]{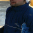} & 
 \includegraphics[width=0.11\textwidth]{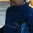} & 
 \includegraphics[width=0.11\textwidth]{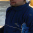} & 
 \includegraphics[width=0.11\textwidth]{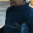} & 
 \includegraphics[width=0.11\textwidth]{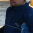} & 
 \includegraphics[width=0.11\textwidth]{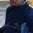} & 
 \includegraphics[width=0.11\textwidth]{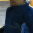} & 
 \includegraphics[width=0.11\textwidth]{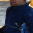}\\


\end{tabular}
\end{figure*}

\begin{figure*}
\tiny
\begin{tabular}{ccccccccc}
Cover & Identity & Dropout & Cropout & Crop & Gaussian & JPEG-mask & JPEG-drop & Combined\\

\includegraphics[width=0.11\textwidth]{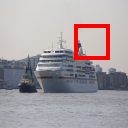} & 
 \includegraphics[width=0.11\textwidth]{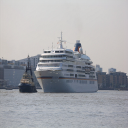} & 
 \includegraphics[width=0.11\textwidth]{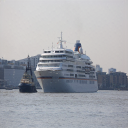} & 
 \includegraphics[width=0.11\textwidth]{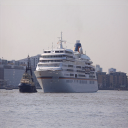} & 
 \includegraphics[width=0.11\textwidth]{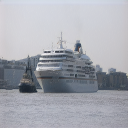} & 
 \includegraphics[width=0.11\textwidth]{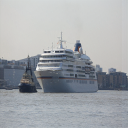} & 
 \includegraphics[width=0.11\textwidth]{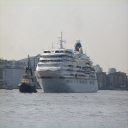} & 
 \includegraphics[width=0.11\textwidth]{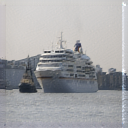} & 
 \includegraphics[width=0.11\textwidth]{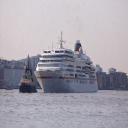}\\

\includegraphics[width=0.11\textwidth]{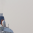} & 
 \includegraphics[width=0.11\textwidth]{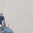} & 
 \includegraphics[width=0.11\textwidth]{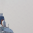} & 
 \includegraphics[width=0.11\textwidth]{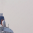} & 
 \includegraphics[width=0.11\textwidth]{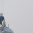} & 
 \includegraphics[width=0.11\textwidth]{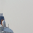} & 
 \includegraphics[width=0.11\textwidth]{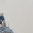} & 
 \includegraphics[width=0.11\textwidth]{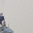} & 
 \includegraphics[width=0.11\textwidth]{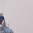}\\


\includegraphics[width=0.11\textwidth]{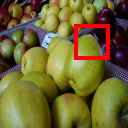} & 
 \includegraphics[width=0.11\textwidth]{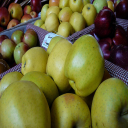} & 
 \includegraphics[width=0.11\textwidth]{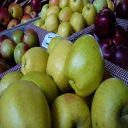} & 
 \includegraphics[width=0.11\textwidth]{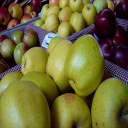} & 
 \includegraphics[width=0.11\textwidth]{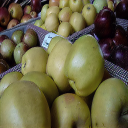} & 
 \includegraphics[width=0.11\textwidth]{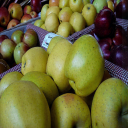} & 
 \includegraphics[width=0.11\textwidth]{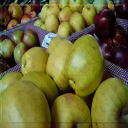} & 
 \includegraphics[width=0.11\textwidth]{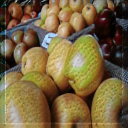} & 
 \includegraphics[width=0.11\textwidth]{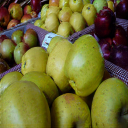}\\

\includegraphics[width=0.11\textwidth]{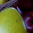} & 
 \includegraphics[width=0.11\textwidth]{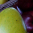} & 
 \includegraphics[width=0.11\textwidth]{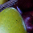} & 
 \includegraphics[width=0.11\textwidth]{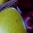} & 
 \includegraphics[width=0.11\textwidth]{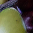} & 
 \includegraphics[width=0.11\textwidth]{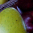} & 
 \includegraphics[width=0.11\textwidth]{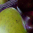} & 
 \includegraphics[width=0.11\textwidth]{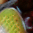} & 
 \includegraphics[width=0.11\textwidth]{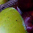}\\


\end{tabular}
\end{figure*}

\endgroup
\end{center}

\end{document}